\documentclass[balance,upint,subscriptcorrection,varvw,mathalfa=cal=boondoxo,spanish,french,vietnamese,russian,greek,pdf-a,colorlinks,nofoot]{asmeconf2}

\titleformat{\section}{\mathversion{sansbold}\bfseries\sffamily\raggedright}{\thesection .}{0.5em}{}

\usepackage{multirow}
\usepackage{graphicx}
\usepackage{float}
\usepackage{stfloats}
\begin{document}




\title{Multi-modal Machine Learning for Vehicle Rating Predictions Using Image, Text, and Parametric Data} 


 
%
%
%

\SetAuthors{%
	Hanqi Su\affil{1} \CorrespondingAuthor{hanqisu@mit.edu},
	Binyang Song\affil{1}, 
	Faez Ahmed\affil{1} 
	}

\SetAffiliation{1}{Massachusetts Institute of Technology, Department of Mechanical Engineering, Cambridge, MA }


\maketitle



\keywords{Multi-modal Learning, Machine Learning, Vehicle Rating Prediction, Model Interpretability, Sensitivity Analytsis}


\begin{abstract}

Accurate vehicle rating prediction can facilitate designing and configuring good vehicles. This prediction allows vehicle designers and manufacturers to optimize and improve their designs in a timely manner, enhance their product performance, and effectively attract consumers. However, most of the existing data-driven methods rely on data from a single mode, e.g., text, image, or parametric data, which results in a limited and incomplete exploration of the available information. These methods lack comprehensive analyses and exploration of data from multiple modes, which probably leads to inaccurate conclusions and hinders progress in this field. To overcome this limitation, we propose a multi-modal learning model for more comprehensive and accurate vehicle rating predictions. Specifically, the model simultaneously learns features from the parametric specifications, text descriptions, and images of vehicles to predict five vehicle rating scores, including the total score, critics score, performance score, safety score, and interior score. We compare the multi-modal learning model to the corresponding unimodal models and find that the multi-modal model's explanatory power is 4\% - 12\% higher than that of the unimodal models. On this basis, we conduct sensitivity analyses using SHAP to interpret our model and provide design and optimization directions to designers and manufacturers. Our study underscores the importance of the data-driven multi-modal learning approach for vehicle design, evaluation, and optimization. We have made the code publicly available at \url{http://decode.mit.edu/projects/vehicleratings/}.

\end{abstract}

\section{ INTRODUCTION}
From the earliest years of their invention, vehicles have stood as a major contributing factor to both everyday consumer life and global economic development. Since the availability of the internet, most consumers research vehicle evaluation scores online and see them as important references for their vehicle purchasing decisions~\cite{Simbolon2020TheIO}. Vehicle evaluation is likewise at the heart of vehicle design, optimization, and improvement. Effective and efficient vehicle evaluation is essential for designers and manufacturers to enhance the appeal of their new models. Extant research has shown promise in exploiting machine learning (ML) and artificial intelligence for vehicle price prediction~\cite{Ponmalar2022ReviewApproaches,9696839,tsagris2022advanced}, vehicle sales prediction~\cite{Xia123ForeXGBoost:XGBoost}, vehicle purchase criteria~\cite{KumarPanda2022CarApproach}, vehicle evaluation~\cite{LiAEvaluation}, and insurance services\cite{Wang2020ScienceDirectLearning}. When evaluating a vehicle, consumers typically analyze multiple data types, such as images, 3D models, parametric specifications, and text reviews. 

Consider a typical vehicle purchasing journey. Initially, a potential vehicle buyer determines the need to purchase a vehicle, which leads them to explore various automotive websites, such as US News, to evaluate numerous vehicle options. In order to make a well-informed choice, they might scrutinize the vehicle's exterior and interior images to assess its design and features. They might even engage with 3D models, when accessible, for a more detailed understanding of the vehicle's attributes.

Additionally, the buyer might review parametric data to measure the vehicle's specifications against others in its category, focusing on elements such as engine capacity, fuel efficiency, safety features, and cost. Reading reviews and written summaries about the vehicle's performance, reliability, and user experience also aids them in ensuring it suits their requirements. Renowned entities, like US News, often rank or rate different vehicles, which can significantly influence the buyer's decision. Through the amalgamation of this varied information, buyers are able to make knowledgeable purchase decisions. Subsequently, they might visit a vehicle dealership to inspect and test drive their chosen vehicle. Upon assessing the vehicle's performance, comfort, and additional features, buyers determine whether it's the right fit for them. If satisfied, they return to the dealership to discuss the price and finalize the purchase. It's crucial to acknowledge that individuals tend to consider multiple data modalities when interacting with designs. However, the majority of current machine learning algorithms are focused on a single modality, typically images, which limits their perspective and hence their practicality. This single-dimensional approach inevitably results in oversimplified conclusions and findings.


This paper endeavors to bridge this gap by tackling the research question: How does multi-modal information about a vehicle influence its ratings? This question is approached utilizing a multi-modal learning method and interpretability models. The application of artificial intelligence and multi-modal deep learning to evaluate and analyze vehicles is relatively unexplored, predominantly due to the substantial requirement for labeled multi-modal data to train deep neural networks. To remedy this shortfall, we also collected a novel multi-modal dataset that includes parametric specifications, images, and textual descriptions of vehicles, all labeled with various vehicle assessment scores.


On this basis, we develop and validate a multi-modal learning model to predict the rating scores of vehicles more comprehensively and accurately. We show that multi-modal learning can exploit the features learned from different types of data and capture the interactions between them to achieve better performance than unimodal learning. Our contributions include the following:

\begin{enumerate}

\item We propose the development of individual unimodal ML models that independently learn from parametric specifications, images, and text descriptions of vehicles. These models aim to predict five distinct vehicle rating scores, namely the total score, critics score, performance score, safety score, and interior score.

\item We introduce a multi-modal learning model capable of concurrently learning from parametric, image, and text data to predict vehicle rating scores. Our findings indicate that this multi-modal learning model markedly outperforms the unimodal models.

\item We assess the relative informativeness of different data modes. Our analysis suggests that parametric data is the most informative for predicting all rating scores, and in most instances, text descriptions offer more predictive power than images.

\item We demonstrate that the sensitivity analyses using SHAP are capable of interpreting our models and providing more detailed design, optimization, and improvement directions to designers and companies.





\end{enumerate}
The rest of this article is organized as follows: Section 2 reviews the approaches to the relevant components of the proposed model. Section 3 introduces the source and composition of the data used in this paper, the data processing module, and both the unimodal and multi-modal machine learning models. Section 4 reports and discusses the performances of the unimodal and multi-modal machine learning models, interprets the models through sensitivity analyses, and summarizes the limitations of this study. Section 5 concludes this paper by highlighting its findings and contributions.

\section{BACKGROUND}
Good vehicle evaluation often requires the analysis of multi-modal data, often involving vehicle parametric specifications, text descriptions, and images. In this section, we first discuss why vehicle evaluation is important. We then review relevant methods for embedding parametric, text, and image data, and investigate prior research on ML techniques for multi-modal data.

\subsection{Why are vehicle evaluations important?}

A few websites provide vehicle reviews and ratings, such as J.D. Power\footnote{\url{https://www.jdpower.com/cars/rankings}}, US News\footnote{\url{https://cars.usnews.com/cars-trucks/rankings}}, Motor Trend\footnote{\url{https://www.motortrend.com/cars/}}, Edmunds\footnote{\url{https://www.edmunds.com/new-car-ratings/}}, and Kelley Blue Book\footnote{\url{https://www.kbb.com/cars/}}. Among these, US News is one of the most popular websites, showing as the number one search result for the query ``vehicle rating'' on search engines such as Google, Bing, and Duck.

US News vehicle ratings are highly influential and are widely followed by consumers who are in the market for a new vehicle. When a vehicle receives high ratings, it can receive increasing consumer interest, ultimately resulting in more sales. US News vehicle ratings consider various factors such as safety, reliability, performance, and interior features. These ratings are based on objective data and evaluations from automotive experts, which can provide consumers with a valuable reference for making informed decisions when purchasing a vehicle. Consumers may use these ratings as a guide when comparing different models and brands and may be more likely to consider a vehicle that has received high ratings. Similarly, vehicle dealerships may use these ratings in their advertising and marketing efforts to attract customers to their inventory. 


Vehicle manufacturers can use US News vehicle ratings to improve their new vehicle designs in several ways:

\begin{enumerate}
    \item Identify Areas for Improvement: By looking at the rating scores for factors like safety, reliability, performance, and interior features, vehicle manufacturers can use these ratings to identify areas where their new vehicles are falling short and improve their designs. US News vehicle ratings take into account consumer needs and preferences. By using the ratings to inform their new vehicle designs, vehicle manufacturers can create vehicles that can better meet the needs and preferences of their target customers.
    \item Benchmark Against Competitors: Vehicle manufacturers can use vehicle ratings to see how their new vehicle designs compare to those of their competitors. This can help them identify areas where they need to improve to remain competitive in the market.
    \item Incorporate Best Practices: Vehicle manufacturers can analyze the highest-ranked vehicles in their category to discover and incorporate best practices into their new vehicle designs. This can help them improve their ratings in future years.
\end{enumerate}

In summary, by using vehicle ratings to inform their new vehicle designs, vehicle manufacturers can create vehicles that better meet the needs of their customers, are more competitive in the market, and ultimately achieve higher ratings in future years.

\paragraph{What are the benefits of predicting vehicle ratings using machine learning?}

Predicting vehicle ratings using ML can be incredibly useful for several reasons. By analyzing a vast amount of data, ML algorithms can identify patterns and correlations across different vehicle data, that may not be immediately apparent to humans. This can help vehicle manufacturers gain valuable insights into the features and characteristics contributing to high ratings. By predicting ratings, vehicle manufacturers can identify areas for improvement in their products and make adjustments to enhance their performance in these areas. Predicting ratings can also help vehicle manufacturers remain competitive in the market by identifying trends and preferences among consumers, allowing them to create products that better meet the needs of their target audience. 

Predicting vehicle ratings can also inform marketing and advertising strategies by highlighting the features that are most important to consumers. Additionally, it can help vehicle manufacturers identify areas for improvement in their vehicle designs and assess their performance relative to their competitors. By tracking their progress over time and setting internal targets for improvement, vehicle manufacturers can use predicted vehicle ratings as a benchmark to inform their product development and competitive strategies. Ultimately, predicting vehicle ratings using ML can help vehicle manufacturers create better products, improve their marketing and advertising strategies, and gain competitive advantages in the market. Next, we discuss different modalities of data in which vehicle information is typically captured.

\subsection{Representing Engineering Data in Different Modalities}

\paragraph{Parametric data}
Engineering product specifications are often provided in the form of tables in a structured way. Parametric data is one of the most commonly used forms of data, consisting of samples (rows) that share the same feature set (columns), which has been used in many applications~\cite{borisov2022deep}. Compared with image or text data, parametric data is mostly heterogeneous, consisting of continuous-valued and categorical-valued attributes. Parametric data features dense values but sparse classification. Although parametric data modeling has been explored intensively using traditional ML methods in the past decades, such as linear regression~\cite{Su2012LinearRegression}, the Gaussian process~\cite{wang2005gaussian}, and gradient-boosted decision trees (GBDT)~\cite{chen2016xgboost}, deep neural networks can learn parametric data in a gradient-based way and allow for the integration of parametric data with other data modalities for multi-modal learning. Typically, parametric data can be learned by simple neural networks, such as multi-layer perceptrons (MLPs). Prior studies have reported that regularization can improve the performance of MPLs in learning parametric data~\cite{kadra2021regularization}. Deep learning techniques like attention mechanisms~\cite{arik2021tabnet} and transformer~\cite{vaswani2017attention} architectures have also been applied to parametric data learning and have shown good prospects.

\paragraph{Image data} With the recent advances of deep learning in computer vision, convolutional neural networks (CNNs) have made breakthroughs in image recognition~\cite{8320684}, image classification~\cite{rawat2017deep}, image segmentation~\cite{minaee2021image}, image generation~\cite{elasri2022image}, and other applications. Therefore, we focus on CNNs for image learning. A few pre-trained image embedding modules are commonly used for image learning tasks, including AlexNet~\cite{krizhevsky2017imagenet}, VGGNet~\cite{simonyan2014very}, ResNet~\cite{he2016deep}, and Inception~\cite{szegedy2015going}. Although current image learning for prediction tasks mostly focuses on classification and recognition, this study particularly focuses on the prediction of vehicle rating scores, which is essentially a regression problem. Different from classification problems, the features learned by the image embedding modules are not used to predict a categorical class through the Softmax activation function but are employed to predict continuous values (i.e., vehicle rating scores) through the Rectified Linear Unit (ReLU) activation function. 

\paragraph{Text data} In addition, natural language processing (NLP) has made significant strides toward automatic comprehension of text data. A few neural network language models (NNLMs)~\cite{bengio2000neural} first appeared to learn massive text data. Then, deep recurrent neural networks (RNNs)~\cite{schmidhuber2015deep} brought NLP to the next level with their strengths in learning sequential data. Its variants, such as long short-term memory (LSTM)~\cite{hochreiter1997long} or gated recurrent unit (GRU)~\cite{chung2014empirical}, were proposed to resolve the problems of gradient vanishing and the explosion of RNNs. Recently, with the advent of the transformer models~\cite{vaswani2017attention}, large Transformer-based language models have gradually gained prominence in fulfilling various NLP tasks. These models have many advantages over the previous NNLMs and RNN-based models: taking entire sequences as input, they can understand the context of each word in a sequence more comprehensively; transformers can process and train more data in less time and utilize the embedded self-attention mechanism to enhance learning. Deep learning models, such as the generative pre-trained transformers (GPT) models~\cite{radford2018improving,radford2019language} proposed and constantly updated by the OpenAI team, and the bidirectional encoder representations from transformers (BERT) model~\cite{devlin2018bert}, and its variants~\cite{liu2019roberta,yang2019xlnet}, significantly improve NLP tasks. In this study, we use a BERT model to encode text data and predict different rating scores.

\subsection{Multi-modal Learning}
On the vehicle rating websites, each vehicle is represented in multiple data modes. Capturing the complementarity and alignment of multi-modal data can lead to a better understanding and more accurate evaluation of a vehicle. Multi-modal learning models that can learn vehicle features simultaneously from the multi-modal information are required to predict vehicle rating scores using such information. In multi-modal learning, the unimodal models are often pre-trained to learn features from each data modality first. On this basis, the multi-modal model can be constructed by fusing the features learned by multiple unimodal models for the downstream tasks. In this paper, we focus on employing multi-modal learning to learn vehicle images, text descriptions, and parametric specifications to predict vehicle rating scores.

Obtaining multi-modal latent representations with effective information fusion lies at the heart of multi-modal learning for this prediction task. Joint representations and coordinated representations are the common options to represent multi-modal data~\cite{Song2023Multi-modalDirections}. Joint representation is better at capturing complementary information from different modalities compared to coordinated representations~\cite{Baltrusaitis2019multimodalTaxonomy, Song2023Multi-modalDirections}, making it more suitable for prediction tasks~\cite{Song2022HEYAnd}. Fusing the information from multiple modalities effectively is critical to learn informative joint representations. Operation-based methods, bilinear pooling methods, and attention-based methods are commonly used for information fusion in multi-modal learning~\cite{Song2023Multi-modalDirections}. The operation-based methods integrate features learned from unimodal data using simple operations, such as concatenation~\cite{Nojavanasghari2016DeepPrediction, Anastasopoulos2019NeuralFeatures, Vielzeuf2019CentralNet:Fusion, Song2022HEYAnd}, averaging~\cite{Shutova2016BlackFeatures}, element-wise multiplication~\cite{Cao2016DeepRetrieval}, (weighted) summation~\cite{Anastasopoulos2019NeuralFeatures, Vielzeuf2019CentralNet:Fusion}, linear combination~\cite{Anastasopoulos2019NeuralFeatures}, and majority voting~\cite{Morvant2014MajorityFusion}. Bilinear pooling fusion integrates features learned from unimodal data by calculating their outer product or Kronecker product~\cite{Zadeh2017TensorAnalysis, Chen2019PathomicPrognosis}. This approach can capture the high-order multiplicative interactions among all modalities, leading to more expressive and predictive multi-modal representations for fine-grained recognition~\cite{Tenenbaum2000SeparatingModels, Zadeh2017TensorAnalysis}. In comparison, the attention-based methods can model dependencies between two data modalities dynamically and assign higher weights to the elements more relevant to the other modality~\cite{vaswani2017attention, Graves2014NeuralMachines}. The integration of the features learned from different modalities can be joined at early or late stages. It is easier to learn the interactions between different data modalities when the features are joined at early stages. However, early joining results in higher-dimensional joint representations, which need more computational resources to train~\cite{Song2023Multi-modalDirections}.

In recent years, multi-modal learning has been explored for a variety of tasks, such as cross-modal synthesis~\cite{Rombach2021High-ResolutionModels, Liu2019Point-VoxelLearning, Nichol2021GLIDE:Models, Kim2021DiffusionCLIP:Manipulation}, multi-modal prediction~\cite{DucTuan2021MultimodalDetection}, and cross-modal information retrieval~\cite{Devlin2015LanguageWorks, Kwon2022EnablingLearning}. However, it is still underexplored in the engineering domain. Recently, Yuan, Mation, and Moghaddam~\cite{Yuan2022LeveragingModel} proposed a multi-modal learning model to capture features from images and text for shoe evaluation. Li et al.~\cite{Li2022AAutoencoder} developed a multi-modal target embedding variational autoencoder model for 2D silhouette-to-3D shape translation. Song et al.~\cite{Song2023ATTENTION-ENHANCEDEVALUATIONS} developed an attention-enhanced multi-modal learning model that learns design sketches and text descriptions simultaneously for design metric prediction.

\subsection{Machine Learning Model Interpretability}
In the realm of engineering, there is a growing emphasis not only on the effectiveness and predictive capabilities of ML models but also on their interpretability~\cite{murdoch2019definitions, molnar2020interpretable, molnar2021interpretable}, which indicates if the reasoning process behind the model predictions can be easily comprehended by humans. The greater the interpretability of a model, the more readily people can understand and trust its predictions.

The rapid advancement of deep learning models has facilitated diverse model-independent explanation techniques. For instance, permutation feature importance~\cite{fisher2019all,rose2019explanatory,ahmed2022product} and Shapley Additive exPlanations (SHAP)~\cite{shrikumar2017learning,sundararajan2020many,NIPS2017_7062} have seen widespread applications. Fisher et al. devised the model class reliance (MCR) approach to facilitate the comprehension of Variable Importance (VI) for unknown models~\cite{fisher2019all}. Within Engineering Design, Ahmed et al.~\cite{ahmed2022product} employed a feature permutation-based technique to interpret the predictions of a graph neural network in predicting product relationships. They found factors such as car make, body type, and segment were important for determining co-consideration relationships.
Mukund Sundararajan and Amir Najmi~\cite{sundararajan2020many} proposed Baseline Shapley (BShap), a technique to explore differences in Shapley value attribution across multiple operations. Shrikumar et al.~\cite{shrikumar2017learning} developed Deep Learning Important FeaTures (DeepLIFT), a method that analyzes the contribution of neurons to input in backpropagation networks to ascertain feature importance. And DeepExplainer, an implementation of Deep SHAP, was developed based on SHAP~\cite{NIPS2017_7062} and DeepLIFT~\cite{shrikumar2017learning}. Additionally, Integrated Gradients~\cite{sundararajan2017axiomatic}, in conjunction with SHAP~\cite{NIPS2017_7062} and SmoothGrad~\cite{smilkov2017smoothgrad}, has given rise to the GradientExplainer, which is a variant of the SHAP Explainer. This variant enables the interpretation of image or text model outputs. In our study, we deploy a SHAP-based approach~\cite{NIPS2017_7062} to interpret the outputs of the image, text, and parametric models.

\section{DATA AND METHOD}
\begin{figure*}[h!]
\centering\includegraphics[width=\linewidth]{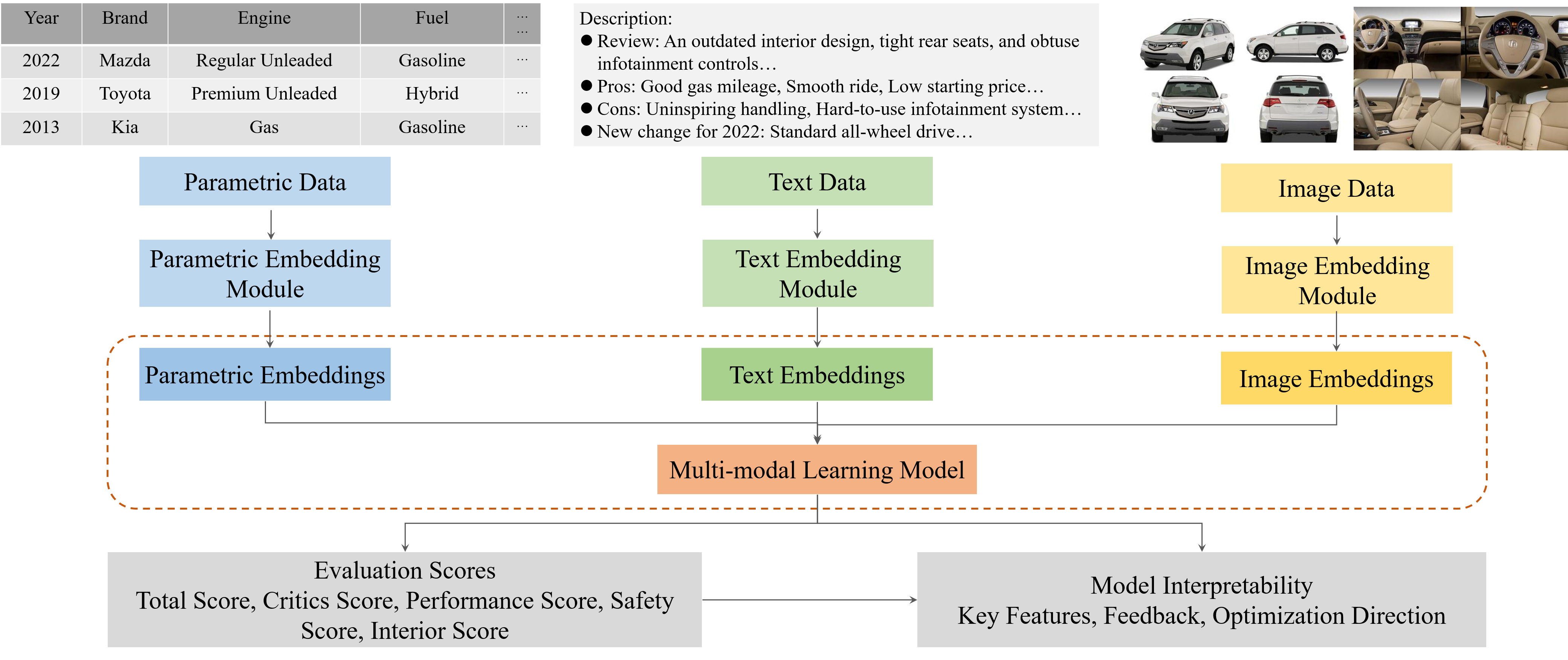}
\caption[Caption for LOF]{The outline of the proposed multi-modal learning model for predicting vehicle rating scores\protect\footnotemark[7].}
\label{fig:1}
\end{figure*}

This section introduces the different types of data we use and the multi-modal learning models for vehicle rating score prediction in this research. This prediction problem is viewed as a regression task. The multi-modal learning model can be divided into five modules, as shown in Figure~\ref{fig:1}. The first module is a pre-processing module to prepare the original data (parametric vehicle specifications, images, and text descriptions). The processed data are the input to the respective unimodal models. The second, third, and fourth modules are the unimodal models capturing features from the parametric, image, or text data, respectively. After the three unimodal models are pre-trained, they can be combined to construct the multi-modal learning model. The rest of this section will separately introduce the data used in this study and each module of the proposed multi-modal learning model.

\subsection{Data}
The data for developing the multi-modal learning model comes from U.S.News\footnote{\url{https://cars.usnews.com/cars-trucks}}. The website provides detailed information on vehicles from different categories, such as sedans, trucks, vans, and sport utility vehicles. The available information covers expert reviews, photos, prices, specifications, performances, rating scores, and so on. In this study, the rating scores, including the overall score, the performance score, the interior score, the critics score, and the safety score, are used as the labels of each vehicle described by the other information. Among them, the performance score reflects the vehicle's performance in terms of acceleration, braking, ride quality, handling, and other qualitative performance metrics. The interior score is regarding vehicle interior manufacturing quality, interior comfort, decoration and features, cargo space, and styling. The critics score represents the reviewer's degree of recommendation and their overall tone regarding the vehicle. The safety score is based on two factors: the number of advanced accident-avoidance technologies and crash test results from the National Highway Traffic Safety Administration and the Insurance Institute for Highway Safety. The overall score for each vehicle is the weighted average of the other four component scores and a few other factors, which are not available on the US News website and are not considered in our study. The rating scores range from 0 to 10, with 10 being the best. 

The goal of our multi-modal regression model is to predict the five scores for new vehicle designs, given their specifications, images, and text descriptions. The text description provides an overall review of the vehicle, including its advantages and disadvantages, and changes to the vehicle model compare to its last version. The image data are the exterior and interior photos of the vehicle. The parametric data conveys detailed specification information of the vehicle, such as body style, dimensions, and other mechanical, safety, and interior features. Our dataset covers 4,517 different vehicle models from different categories from 2007 to 2022. However, some relevant information is missing on the US News website for 1,946 vehicle models, which are excluded from this study. Accordingly, the data of 2,571 vehicles is used to develop the multi-modal learning model.

\footnotetext[7]{\url{https://cars.usnews.com/cars-trucks/acura/mdx/2007}} 

\subsection{Data Processing}
The raw data from the US News website contains three types of data: parametric specifications, text descriptions, and image data. The parametric specifications consist of five categories: general information, exterior information, interior information, mechanical information, and safety information. Each category covers multiple subcategories, as listed in Table~\ref{table:1}. Notably, the subcategories further comprise varying numbers of features, resulting in a total of 302 features for each vehicle. Some of these features are numeric, while others are categorical. When preprocessing the data, we normalize all numeric features to [0,1] and use one-hot encoding to represent the categorical features.

\begin{table}[h!]
    \begin{center}
\begin{tabular}{|c|c|c}
\hline
Category & Subcategory\\
\hline
\multirow{5}{5em}{General Information} & Years \\
& Brand \\
& Drivetrain \\
& Manufacturer Suggested Retail Price (MSRP) \\
& Mile Per Gallon (MPG) City \\
& Mile Per Gallon (MPG) Highway \\
\hline
\multirow{3}{5em}{Exterior Information} & Exterior Body Style \\
& Exterior Dimensions \\
& Exterior Measurements \\
\hline
\multirow{6}{5em}{Interior Information} & Interior Convenience \& Comfort \\
& Interior Dimensions \\
& Interior Entertainment \\
& Interior Heating Cooling \\
& Interior Navigation \& Communication \\
& Interior Seats \\
\hline
\multirow{3}{5em}{Mechanical Information} & Mechanical Transmission \\
& Mechanical Fuel \\
& Engine \& Performance \\
\hline
\multirow{3}{5em}{Safety Information} & Safety Airbags \\
& Safety Brakes \\
& Safety Features \\
\hline
\end{tabular}
\end{center}
    \caption{Parametric specification information.}
    \label{table:1}
\end{table}

\begin{figure}[htbp]
\centering\includegraphics[width=\linewidth]{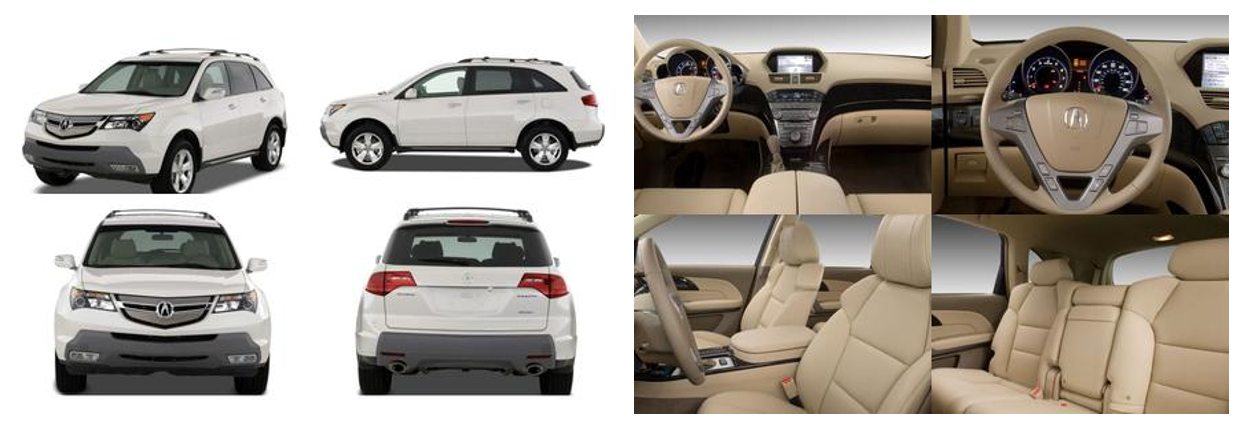}
\caption[Caption for LOF]{Examples of vehicle exterior\protect\footnotemark[8] and interior\protect\footnotemark[9] photos.}
\label{fig:0}
\end{figure}

The image data consists of exterior and interior photos. Among a large number of exterior and interior photos, we select the four most representative exterior or interior photos as the input to the image model. Specifically, the selected exterior photos include photos taken from four fixed views: angular front, front, rear, and side. The original size of these photos is $776 \times 776$. First, we resize the original images to $224 \times 224$. Second, we remove part of the white background at the periphery of the exterior photos to further reduce the size of the images. Third, we integrate the four resized exterior photos into a single exterior image with a size of $448 \times 290$, as shown in Figure~\ref{fig:0}. The selected interior photos cover the major interior components, including the dashboard, front seat, rear seat, and steering wheel. While their original size is $776 \times 517$, we resize the interior photos to $224 \times 150$ proportionally and integrate them to produce a single interior photo with a size of $448 \times 300$.

For text data, the information of different features is formatted as ``The name of this feature: content'', and the information on different features is concatenated successively into a single machine-readable string. The maximum, minimum, and average word lengths for the text descriptions are 224, 32, and 74, respectively. 

\footnotetext[8]{\url{https://cars.usnews.com/cars-trucks/acura/mdx/2007/photos-exterior}} 
\footnotetext[9]{\url{https://cars.usnews.com/cars-trucks/acura/mdx/2007/photos-interior}}
\subsection{Models}
In this subsection, we first introduce the three unimodal models for embedding the parametric data, image data, and text data, respectively. Then, we describe how the multi-modal learning model is constructed based on these three unimodal models.

\begin{figure}[htbp]
\centering\includegraphics[width=\linewidth]{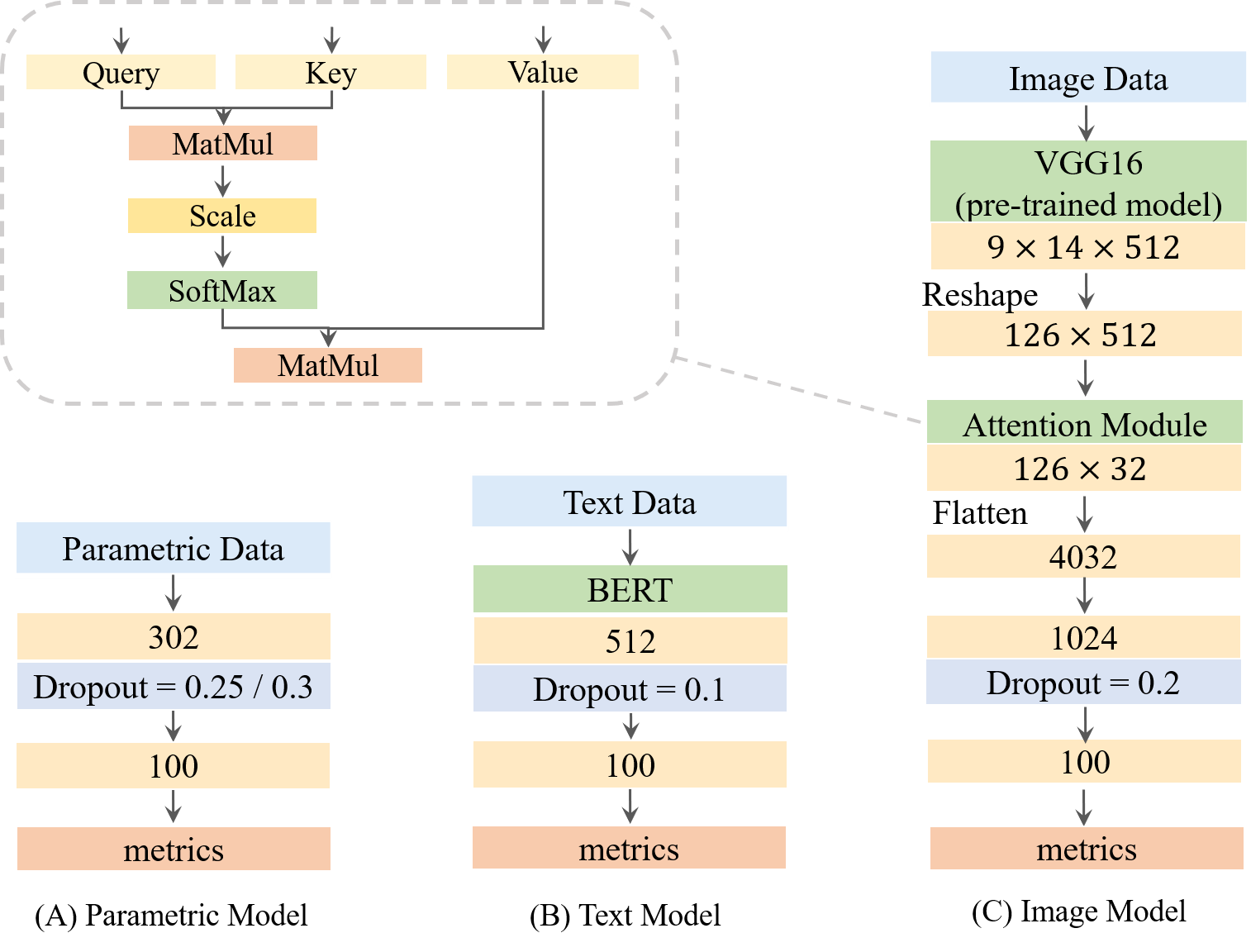}
\caption{The architectures of three unimodal models.}
\label{fig:2}
\end{figure}

\textbf{Unimodal Model} Figure~\ref{fig:2} illustrates the architectures of the unimodal models that respectively learn the parametric, image, and text data. All these unimodal models adopt the ReLU activation function for the regression task in this study. The final output for each unimodal model is the predicted value of the rating score.

(1) \textbf{Parametric model}: Firstly, we construct an MLP model to learn the parametric data. To find the optimal neural network architecture and hyper-parameters, we conduct a set of pilot experiments. This process leads us to a simple neural network architecture containing two hidden layers, as depicted in Figure~\ref{fig:2}-A. The number of neurons in the first hidden layer is equal to the dimension of the input data (302), and that of the second hidden layer is 100. We add a dropout layer after the first hidden layer with dropout rates ranging from 0.25 or 0.3 for predicting different rating scores. 

(2)\textbf{Text model}: Secondly, the text model adopts a pre-trained transformer-based BERT~\cite{devlin2018bert} text embedding module. We use the pooled output from the BERT model as the final embedding of the input text with a dimension of 512. During pre-training, the BERT embedding module is trained on large text databases (e.g., Wikipedia.) for multiple tasks. The adoption of the pre-trained BERT model allows for effective knowledge transfer from the large external text dataset to our target text descriptions when we fine-tune the model with our dataset for the regression task. Following the BERT embedding layer, a dropout layer with a dropout rate of 0.1 and a dense layer with 100 neurons are attached before the final output layer, as shown in Figure~\ref{fig:2}-B. We unfreeze all layers to train the text model.

(3) \textbf{Image model}: Thirdly, we construct a CNN-based model to learn the vehicle images. We experiment with multiple CNN models, including ResNet~\cite{he2016deep}, Inception~\cite{szegedy2015going}, and VGG16~\cite{simonyan2014very}, during our pilot experiments and get similar performances from them. VGG16~\cite{simonyan2014very} is selected for this study because it takes less time to train. The output from the VGG16 embedding module exhibits a dimension of $9\times14\times512$. Following the image embedding module, we add a self-attention mechanism, as visualized in Figure~\ref{fig:2}-C. It reshapes the output to $126\times512$, which is seen as a set of 126 latent features with a dimension of 512. The self-attention mechanism employs a latent dimension of $32$ in this study. Since the input to the image model integrates four exterior or interior vehicle photos, which complement or align with each other to different degrees. The self-attention mechanism is expected to facilitate capturing the interactions between different regions of the input images. The self-attention mechanism employs the dot-product attention proposed in ``Attention Is All You Need''~\cite{vaswani2017attention}. After that, a flatten layer, a dense layer with 1024 neurons, a dropout layer with a dropout rate of 0.2, another dense layer with 100 neurons, and a final output layer are attached sequentially.

For the image model, to enhance our evaluation of vehicle features, we use interior photos to evaluate the interior score and use exterior images to evaluate the other rating scores.

\begin{figure}[htbp]
\centering\includegraphics[width=\linewidth]{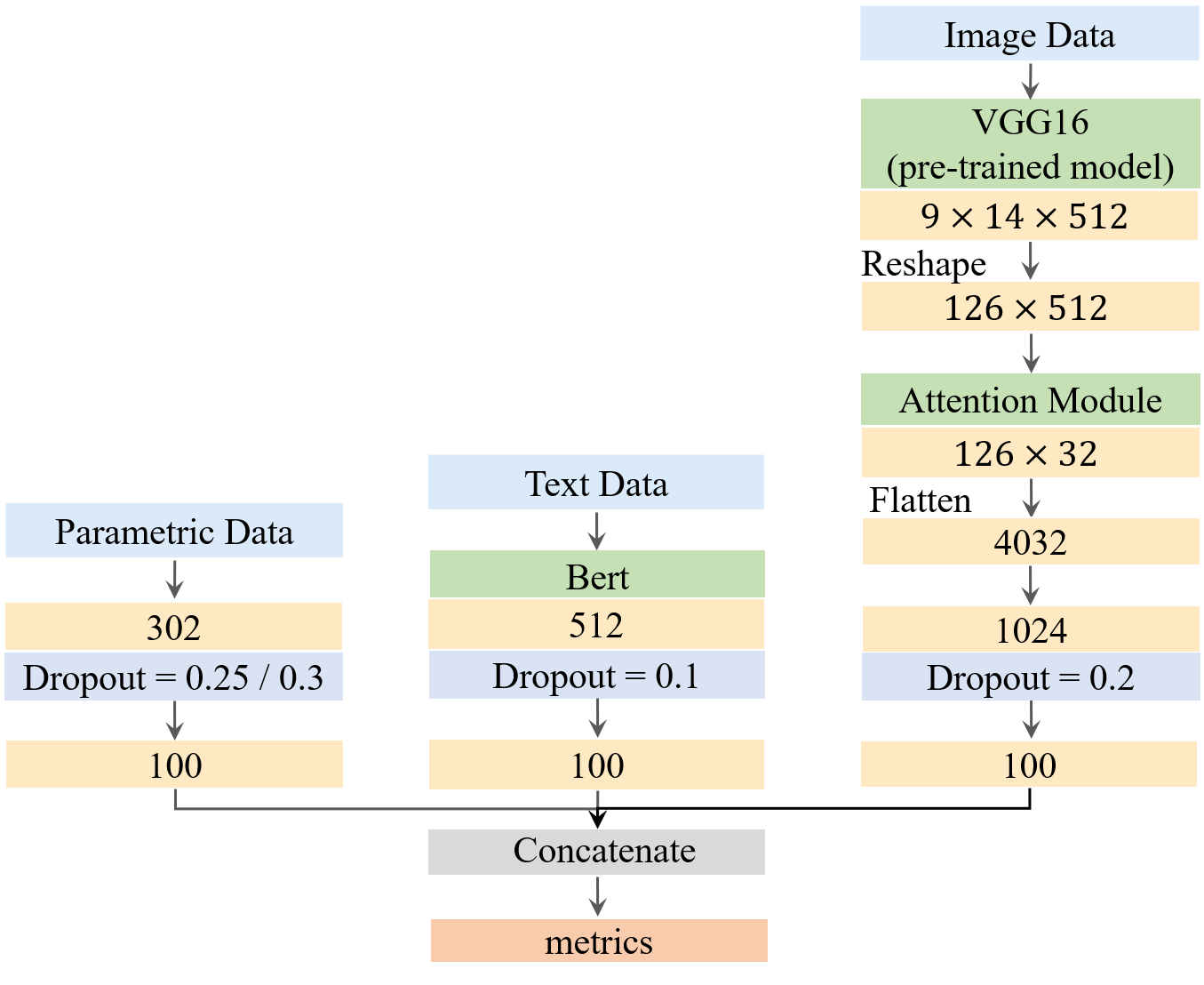}
\caption{The architecture of the multi-modal learning model.}
\label{fig:3}
\end{figure}

\textbf{Multi-modal Learning Model} After the three unimodal models are trained, we integrate them to construct the multi-modal learning model. To facilitate the learning of the interactions between the three data modalities, we do not directly integrate the unimodal models by concatenating their final outputs. Instead, we concatenate the final embedding of the input parametric, text, and image data from the corresponding unimodal models as the final joint representation of the multi-modal input data. The final output of the multi-modal learning model is calculated from the joint representation through a dense layer with the ReLU activation function. The architecture of the multi-modal model is shown in Figure~\ref{fig:3}. In comparison, we also construct three bi-modal models that respectively combine two of the three unimodal models. These different combinations give us four multi-modal learning models. For the sake of simplicity, we refer to the bi-modal learning model combining parametric  and text data as $Par\_Text-MML$ model, the bi-modal learning model combining parametric and image data as $Par\_Img-MML$ model, the bi-modal learning model combining the image and text data as $Img\_Text-MML$ model. The multi-modal model combining the parametric, text, and image data is called $Par\_Text\_Img-MML$ model in this paper hereafter. When training the multi-modal models, we initialize the multi-modal learning models with the pre-trained weights from the unimodal models and fine-tune the weights jointly to learn the interactions between the three data modalities for better vehicle rating score prediction.

\section{RESULTS AND DISCUSSION}
In this section, we compare the performances of the three unimodal models and the four multi-modal learning models to verify the effectiveness of the proposed multi-modal learning model. Specifically, the performance of each model is assessed in terms of the explanatory power for the variances of the vehicle rating scores, which is known as the determination coefficient (i.e., $R^2$ value) in statistics. In regression, the degree of fit improves as the $R^2$ value increases. To train and test the model, the 2,571 vehicles in our dataset are divided into the training set, validation set, and test set following the ratio of 0.8:0.1:0.1. In the process of data split, we ensure that the stratified distribution of the rating scores within each set is consistent with that of the entire dataset. We observe that the distribution of different rating scores could be very different, so we generate a unique data split for each of the five rating scores. All the models use the same data split to predict the same rating score for easy comparison. During training, different unimodal and multi-modal learning models are trained with the same batch size of $32$ and the initial learning rates ranging from $0.001$ to $0.00005$, which are selected through a series of pilot experiments. We also apply different decay rates ranging from $0.0$ to $e^{-0.015}$ to schedule the learning rates during training different models. The training process is ended if the validation loss does not decrease for 20 consecutive epochs. In order to demonstrate the stability of the model and test the statistical significance of the differences between different models, we repeat each experiment 10 times.

\subsection{Performance of Unimodal Models}
\begin{figure}[htbp]
\centering\includegraphics[width=\linewidth]{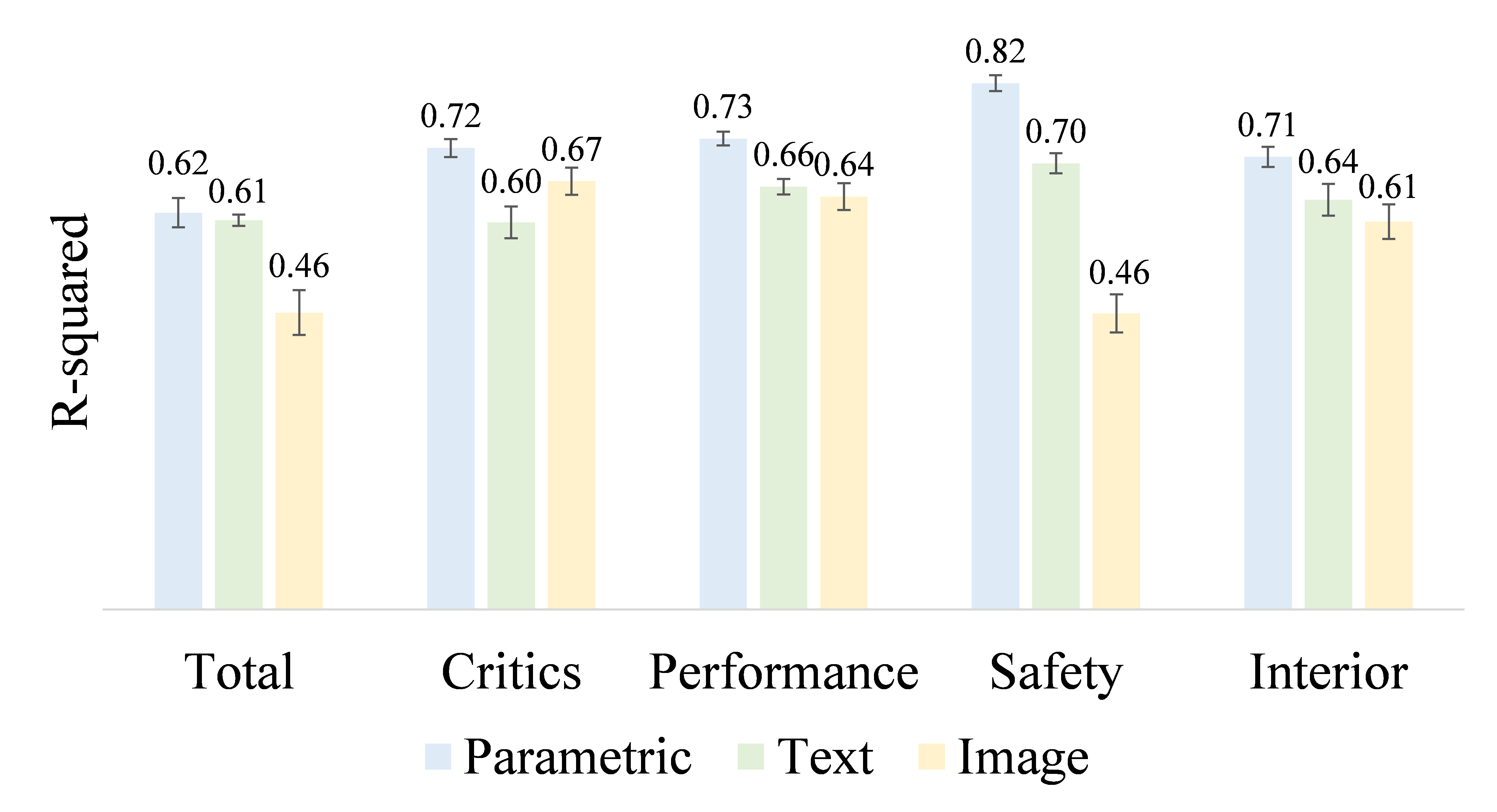}
\caption{The performances of the unimodal models. The columns show the average $R^2$ values from the 10 repeated experiments with the bars indicating one standard error. We observe that parametric models have higher $R^2$ across different metrics.}
\label{fig:4}
\end{figure}

The three unimodal models show varying performances in predicting different rating scores, as shown in Figure~\ref{fig:4}. The parametric model best predicts all rating scores. Its $R^2$ values are at least 0.04 higher than that of the corresponding image and text models for predicting all rating scores. Moreover, in most cases, the text model outperforms the image model. That is, the parametric data is most informative while the image data is least informative in predicting these rating scores. The information conveyed by these different types of data may explain the differences in model performance. The parametric data intuitively shows the detailed specifications of a vehicle, including general information, exterior information, interior information, mechanical information, and safety information of the vehicle, which summarizes the vehicle's major characteristics. The compact representation may make it easier for the model to capture the key features, leading to better predictions. In comparison, the text data describes the advantages and disadvantages of a vehicle compared to other vehicles or its previous version, which is also valuable for rating the vehicle. The images of a vehicle show its aesthetic features and body design, which influence customers' affection for it and its aerodynamic performance. Since exterior design is not considered by the five rating scores directly, the information conveyed by the images might be less informative for predicting these scores.

Moreover, we find that the three unimodal models are relatively less effective in predicting the total score compared to the other scores. The total score is an overarching evaluation of a vehicle, which is the weighted average of the other four rating scores and several other indicators. The prediction of such an overarching score needs more complicated and comprehensive information from multiple perspectives, which is more challenging for the unimodal models to learn from a single data modality. Therefore, the unimodal models may struggle to learn enough features during training and thus do not predict the overall score as well as the other four rating scores. In addition, the dataset used in this study is small, which cannot provide sufficient information to train these large models. We observe that it is easy to overfit these models and the training is terminated early before better model weights can be learned, which may lead to insufficient final predictions. Among all five rating scores, the parametric model exhibits the highest $R^2$ value in predicting the safety score, and the $R^2$ values of the three unimodal models differ greatly. The $R^2$ value of the parametric model is higher than that of the worst model (image) by 0.34. One potential reason is that its evaluation is partly based on the advanced accident avoidance technologies implemented by a vehicle, which is described clearly in the parametric data. In comparison, although the vehicle body design reflected by the images affects the safety of the vehicle, the material of the vehicle body is unknown from the images and the importance of body design is being weakened by the incorporation of the technologies in recent years.


\subsection{Effect of Multi-modal learning}
\begin{figure*}[bp]
\centering\includegraphics[width=\linewidth]{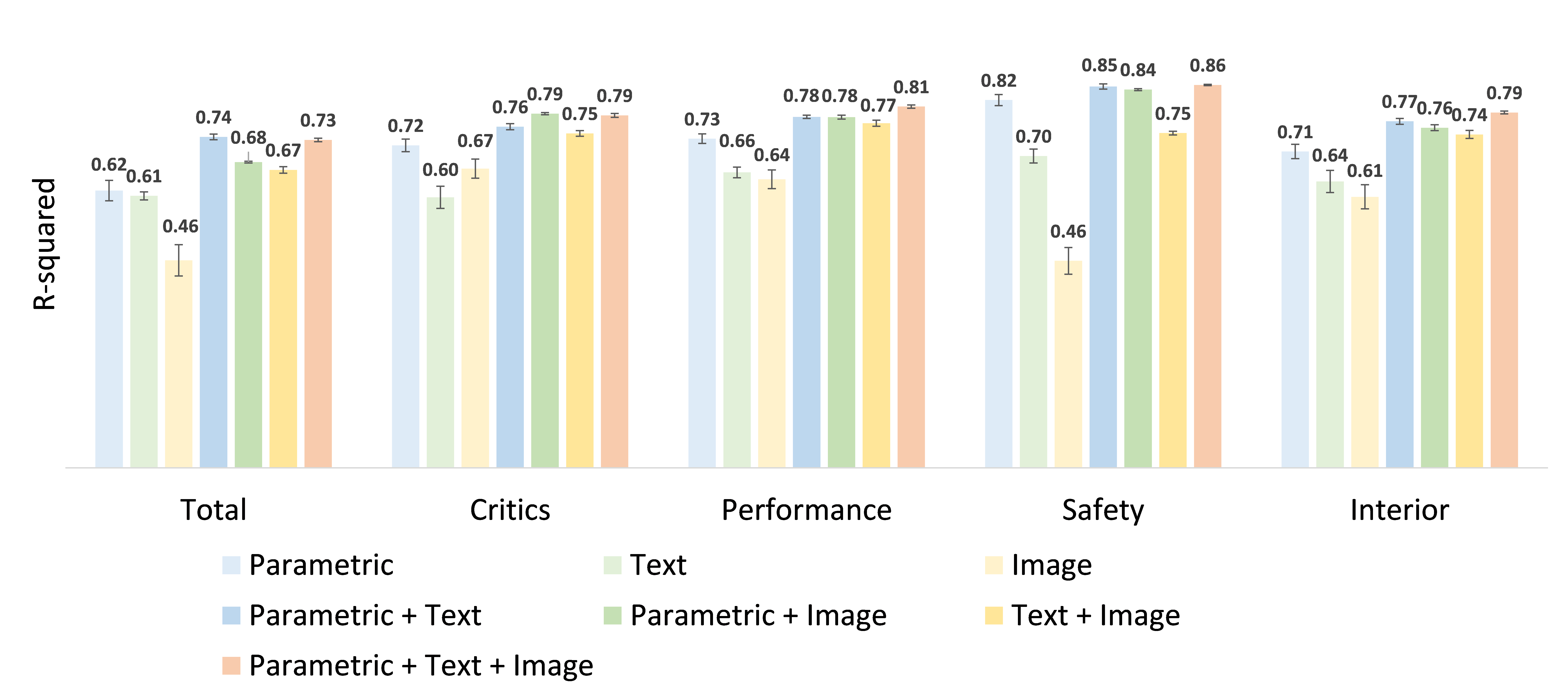}
\caption{The comparison in performance among the unimodal and multi-modal learning models. The columns show the average $R^2$ values with one standard deviation bar. We observe that the multi-modal model using three modalities outperforms unimodal or bi-modal models.}
\label{fig:5}
\end{figure*}

\paragraph{Multi-modal learning models outperform the unimodal models.}
For predicting all rating scores, the average $R^2$ values of the multi-modal learning models are significantly higher than that of the corresponding unimodal models, as shown in Figure~\ref{fig:5}. The results suggest that compared to the unimodal models, the joint learning of multi-modal data enables the multi-modal learning models to leverage the complementary features learned from different modalities to better predict the rating scores. Moreover, the $Par\_Text\_Img-MML$ model also performs better than the three bi-modal learning models that integrate two types of data for predicting all rating scores except for the total score. The $Par\_Text-MML$ model slightly outperforms the $Par\_Text\_Img-MML$ model for predicting the total score. This may seem counterintuitive, as adding one more mode should logically allow the model to learn more information and, thus, likely make better predictions. However, as discussed above, the evaluation of the total score relies on more complicated, interrelated, and comprehensive information. This is more challenging for the models to learn. The features learned from the three data modalities are fused through simple concatenation in this paper, which may not be able to capture the complex interactions among the modalities for better total score prediction when the image data is involved. Another possible reason is the dataset used in this study is not large enough to support learning the complex cross-modal interactions for predicting the total score.

\paragraph{The effect of multi-modal learning varies in predicting different rating scores.}
The effect of multi-modal learning is most substantial in predicting the total score. The best multi-modal model ($Par\_Text-MML$) outperforms the best unimodal model (parametric) by 0.12. The characteristics of the total score imply that its evaluation relies more on a comprehensive understanding of a vehicle. Accordingly, the multi-modal features learned by the multi-modal models help significantly in this regard. The effect of multi-modal learning is least obvious in predicting the safety score. The best $Par\_Text\_Img-MML$ model only improves the $R^2$ value by 0.02 compared with the best unimodal model, which is the parametric model. As mentioned above, the parametric data clearly describe the advanced accident avoidance technologies implemented by a vehicle, which inform the evaluation of the safety score. Since the $R^2$ values achieved by the image model and the text model are much lower than the parametric model, the incorporation of the text and image data only slightly complements the parametric information for this evaluation. In general, the $Par\_Text\_Img-MML$ model exhibits similar or slightly better performances compared to the best bi-modal learning models for predicting all rating scores. The simple information fusion mechanism and the small size of the dataset in this study may help explain this.

\subsection{Implications for Engineering Design}
As demonstrated in the last subsection, our multi-modal learning models can predict vehicle rating scores accurately using vehicle parametric specifications, text descriptions, and images. However, a more detailed interpretation of the outputs from the models is needed to inform designers and companies about potential directions to optimizing the inferior design or advertising the superior design of a vehicle. For this purpose, we utilize the SHAP~\cite{NIPS2017_7062} method to interpret the outputs from the image, text, and parametric models. Through backward gradient-based sensitivity analysis, the output SHAP values indicate the influence of each element of the input data on the final prediction made by a model. A higher absolute SHAP score suggests a higher influence. Therefore, the SHAP method can help us interpret how a deep learning model makes its decision.

\begin{figure}[h!]
\centering\includegraphics[width=\linewidth]{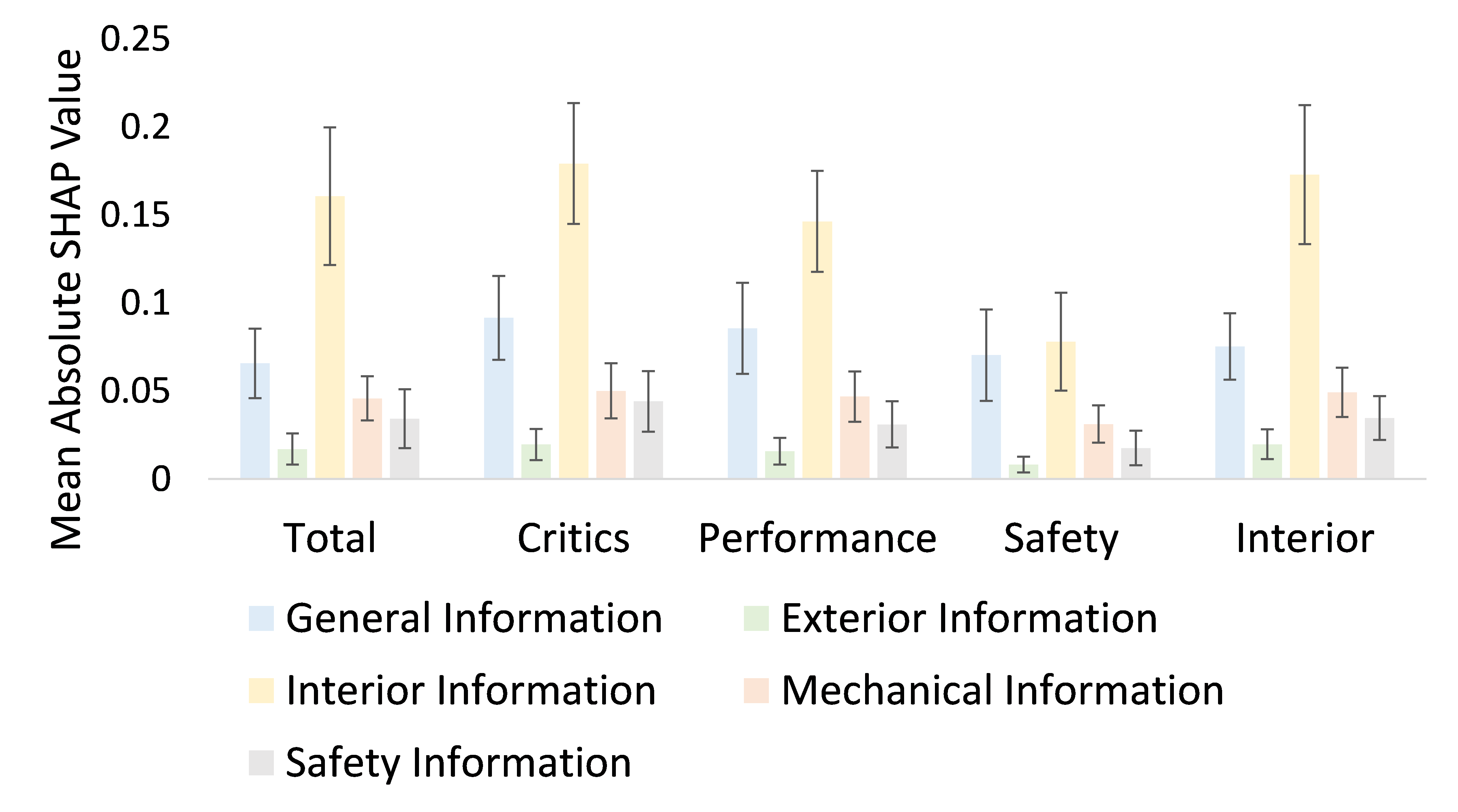}
\caption{Mean Absolute SHAP values of the five feature categories of the parametric data with one standard error. We observe that the interior information category exhibits the largest SHAP values while the exterior information category holds minimal significance.}
\label{fig:6}
\end{figure}

We first conduct SHAP analysis for the parametric model. As we mentioned before, the parametric data conveys rich information across five feature categories as listed in Table~\ref{table:1}. The SHAP analysis can help us identify the most informative and influential vehicle feature categories from the parametric data. We run SHAP analysis for the parametric models that predict the five rating scores, respectively. Figure~\ref{fig:6} illustrates the average absolute SHAP values of the five feature categories across all vehicles in the test set for predicting different scores. These values indicate the extents to which different feature categories affect the model's predictions. The findings indicate that the impact of each category varies, with the interior information category having the greatest influence on all score predictions and the exterior information category having the least impact on them.

On this basis, we further analyze the influences of the 21 feature subcategories on the model predictions, and Table~\ref{table:3} shows the 21 subcategories used to predict the total score. The findings are in line with our expectations and quite interesting. When buying a vehicle, customers tend to base their decisions on the vehicle brand and the appearance of the vehicle's body. For instance, some people prefer Toyota sedans, while others may prefer Subaru SUVs. This indicates that the features like ``Brands'' and ``Exterior Body Style'' can significantly influence the prediction. Furthermore, the comfort and convenience of driving a vehicle are crucial since the most straightforward feeling that drivers and passengers may have for a vehicle is how comfortable and convenient it is to drive or ride in it. That is why most of the interior information subcategories have prominent impacts on rating predictions. Additionally, people value the safety and performance of a vehicle since if the vehicle's safety and performance are not up to par, people are less likely to trust and purchase it. Accordingly, ``Safety Features'', ``Engine \& Performance'' and ``Mechanical Transmission'' features are also important. Designers and companies need to focus on these feature subcategories to improve or advertise their designs.

\begin{table*}[h!]
    \begin{center}
\scalebox{0.9}{
\begin{tabular}{lcl}
\hline
\textbf{Subcategory} & \textbf{Mean Absolute Shap Values} & \textbf{Sample Features}\\
\hline
Interior Convenience \& Comfort & $9.11\cdot10^{-2}$ & Auxiliary Pwr Outlet, HID headlights, Back-Up Camera \\
Brand & $3.93\cdot10^{-2}$ & Toyota, Honda, Ford \\
Interior Seats & $2.59\cdot10^{-2}$ & Heated Rear Seat, Heated Front Seats, Driver Lumbar\\
Interior Entertainment & $2.58\cdot10^{-2}$ & WiFi Hotspot, Smart Device Integration, Entertainment System \\
Engine \& Performance & $2.17\cdot10^{-2}$ & Engine Type, Premium Unleaded, Regular Unleaded \\
Safety Features & $2.06\cdot10^{-2}$ & Lane Keeping Assist, Blind Spot Monitor \\
Mechanical Transmission & $2.01\cdot10^{-2}$ & Continuously Variable Trans, Manual, 6-Speed Automatic \\
Years & $1.78\cdot10^{-2}$ & 2010, 2014, 2020 \\
Exterior Body Style & $1.25\cdot10^{-2}$ & SUV, Sedan, Hatchback \\
Interior Heating Cooling & $7.44\cdot10^{-3}$ & Climate Control, Dual Zone A/C, Rear A/C \\
Safety Airbags & $7.21\cdot10^{-3}$ & Rear Side Air Bag, Passenger Air Bag On/Off Switch \\
Interior Navigation \& Communication & $7.15\cdot10^{-3}$ & Navigation System, Onboard Communications System \\
Safety Brakes & $6.48\cdot10^{-3}$ & Front Disc/Rear Drum Brakes, 4-Wheel Disc Brakes \\
Drivetrain & $6.21\cdot10^{-3}$ & AWD, FWD, RWD \\
Mechanical Fuel & $3.99\cdot10^{-3}$ & Hybrid Fuel, Plug-In Electric/Gas, Gasoline Fuel \\
Exterior Dimensions & $3.43\cdot10^{-3}$ & Dim Width (in), Wheelbase (in), Dim Length (in) \\
Interior Dimensions & $3.30\cdot10^{-3}$ & Front Head Room (in.), Front Leg Room (in.) \\
Manufacturer Suggested Retail Price(MSRP) & $1.46\cdot10^{-3}$ & MSRP \\
Exterior Measurements & $1.08\cdot10^{-3}$ & Base Curb Weight (lbs) \\
Mile Per Gallon (MPG) City & $5.25\cdot10^{-4}$ & MPG City \\
Mile Per Gallon (MPG) Highway & $4.53\cdot10^{-4}$ & MPG Highway \\
\hline
\end{tabular}}
\end{center}
    \caption{The feature subcategories with the corresponding SHAP Values and a few sample features. We observe that Interior convenience and brand are found most important in ratings prediction.}
    \label{table:3}
\end{table*}

\begin{figure}[h!]
\centering\includegraphics[width=\linewidth]{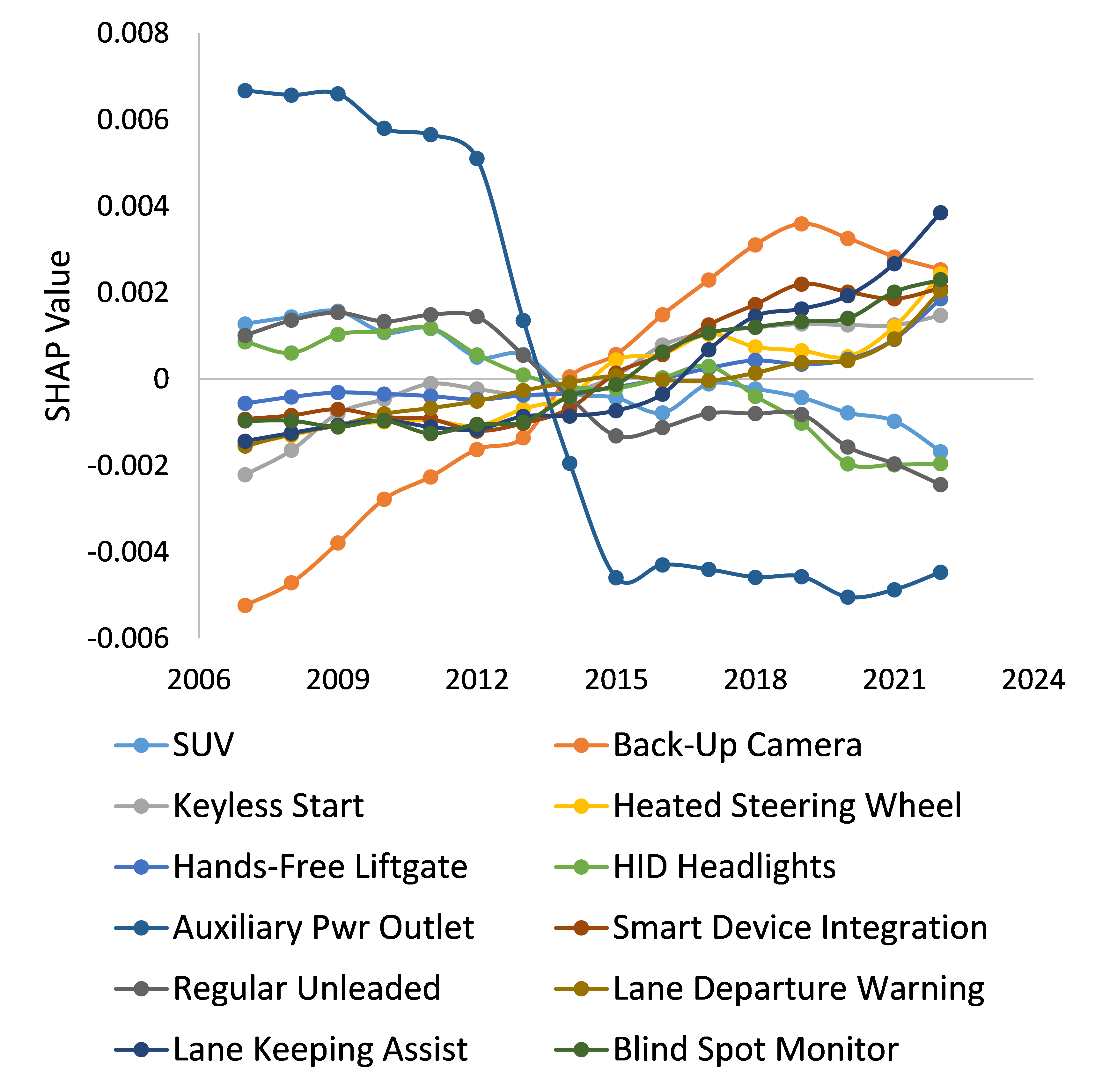}
\caption{The SHAP values of 12 example features for predicting the total score over time. We observe that the importance of some features such as back-up cameras has increased in the last decade.}
\label{fig:8}
\end{figure}

In addition to examining the average absolute values, we also analyze the variation of the average SHAP values for the 302 individual features over time. Although the SHAP values of the majority of the features do not show noticeable trends (e.g., sharp fluctuations or little changes), a subset of the features exhibit clear increasing or decreasing trends, as shown in  Figure~\ref{fig:8}. As electronic and information technologies continue to advance, these technologies have been enhancing the driving experience and promoting driving safety. For example, the ``Heated Steering Wheel'' prevents hand stiffness during long driving hours in winter, and ``Keyless Start'' eliminates the need for manual key insertion by pressing a button inside the vehicle, or turning a knob, making the process more convenient. Furthermore, the ``Hands-Free Liftgate'' automatically opens and closes the liftgate. Other technologies, such as ``Back-UP Camera,'' ``Lane Keeping Assist,'' and ``Lane Departure Warning'' help improve driving safety on the road. These features have experienced an increase in their SHAP values over time, highlighting their growing positive influence on the model's prediction. On the other hand, the SHAP values of a few others show the opposite pattern, such as ``auxiliary power outlet'', ``regular unleaded (fuel)'', and ``high-intensity discharge (HID) headlights''. These features play increasingly negative roles in affecting the predictions, which means having these features may result in lower rating scores as time passes by. For example, ``Auxiliary Pwr Outlet,'' also known as the ``car cigarette lighter,'' is an outdated feature that has been excluded by many new vehicle models. By analyzing the original data, we observe that most vehicles manufactured before 2014 were equipped with cigarette lighters, but very few vehicles had them after that year. Similarly, ``HID headlights'' were once popular for their high brightness and long service life compared to traditional halogen bulbs. However, due to their expensive manufacturing costs and slow response time to peak brightness, they have been gradually replaced by LED headlights that offer lower power consumption, longer lifespan, and faster response times. Consequently, HID headlights are disappearing from the market, which aligns with the observed changes in its SHAP values.

\begin{figure}[h!]
\centering\includegraphics[width=\linewidth]{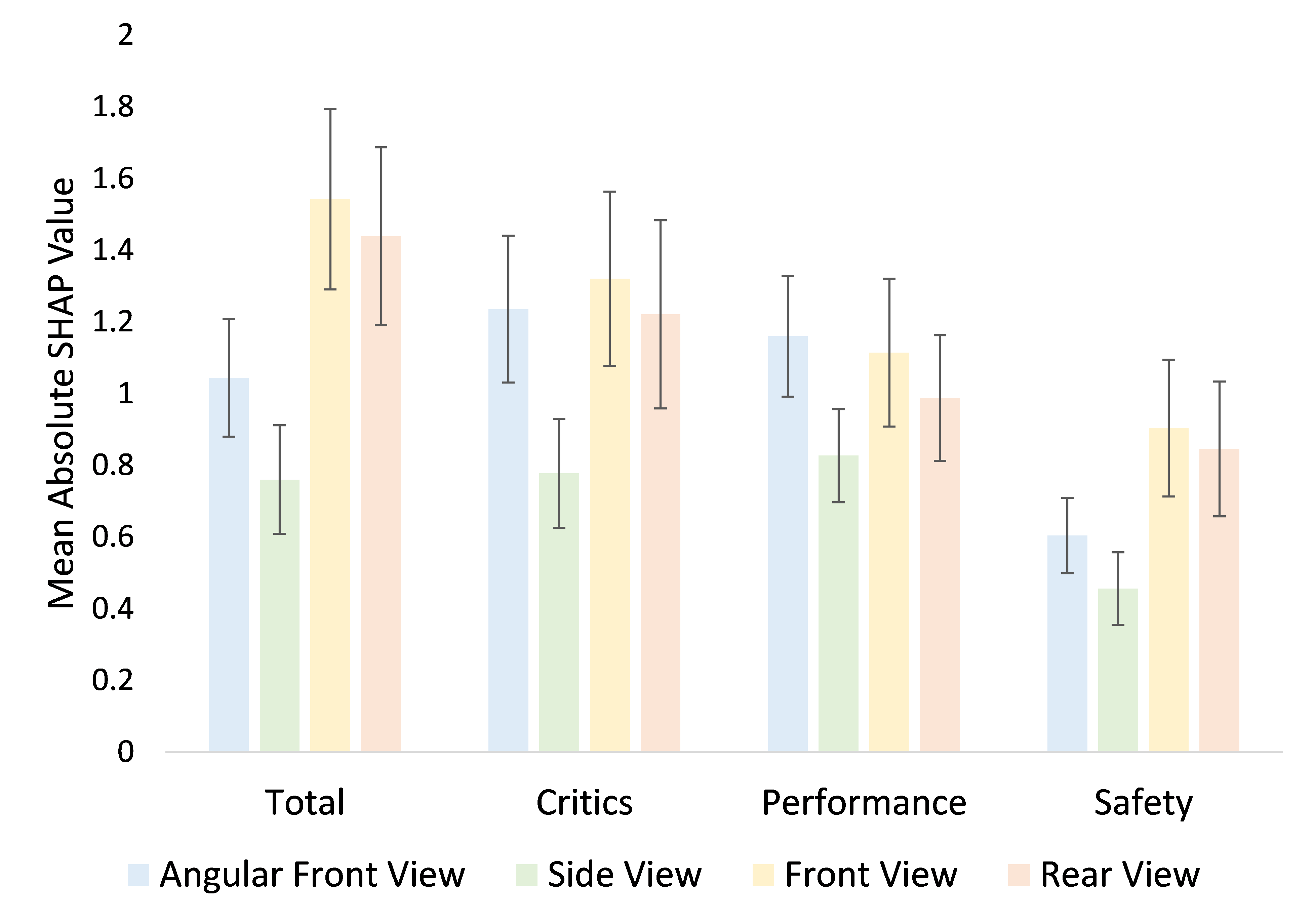}
\caption{Mean absolute SHAP values of the exterior image regions for predicting the total, critics, performance, and safety scores with one standard deviation bar. The SHAP values of the front view and rear view play a bigger role in predicting these scores. }
\label{fig:9}
\end{figure}

Then, We employ the SHAP method to analyze the informativeness of different image regions for predicting different rating scores. Since we use the interior and exterior images to predict the interior score and the other four scores, respectively. The mean absolute SHAP values of interior image regions are displayed in Table~\ref{table:2}, while Figure~\ref{fig:9} showcases the mean absolute SHAP values of the exterior image regions for predicting the four rating scores. For predicting the interior score, the dashboard and steering wheel regions present higher mean absolute SHAP values than the front and rear seat regions, suggesting that the model may primarily rely on features in these regions to predict the interior score. Notably, the SHAP values of the front and rear views in the exterior images are higher than that of the other two views in predicting most of the rating scores. The results indicate the critical role of the front and rear views in predicting the other scores. This indicates that when purchasing a vehicle, people prioritize the front and back views, as they are the most visible when driving. Well-designed front and rear portions of a vehicle can also potentially reduce safety risks.

\begin{table}[h!]
    \begin{center}
\scalebox{0.9}{
\begin{tabular}{|c|c|c|c|c|c}
\hline
Score & Dashboard & Steering Wheel & Front Seat & Rear Seat\\
\hline
Interior & 1.159 & 1.011 & 0.448 & 0.389 \\
\hline
\end{tabular}}
\end{center}
    \caption{Mean absolute SHAP values of the interior image regions for predicting the interior score. The SHAP values of the dashboard and steering wheel highlight their high importance in predicting the interior score}
    \label{table:2}
\end{table}

\begin{figure}[h!]
\centering\includegraphics[width=\linewidth]{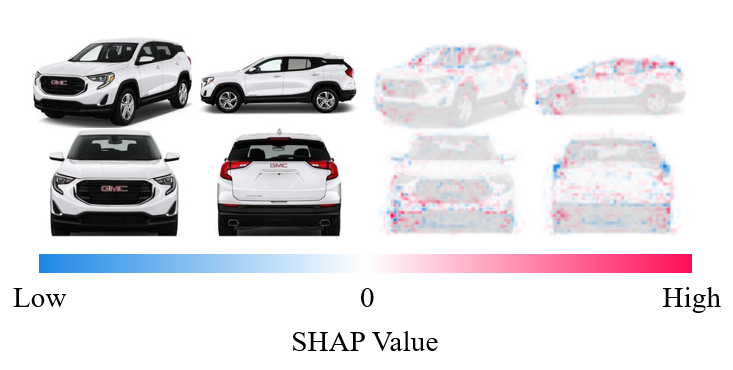}
\caption[Caption for LOF]{SHAP values of the exterior image regions of the 2020 GMC Terrain\protect\footnotemark[10] for total score prediction: the SHAP values on the right with the corresponding exterior image on the left.}
\label{fig:10}
\end{figure}

\begin{figure}[h!]
\centering\includegraphics[width=\linewidth]{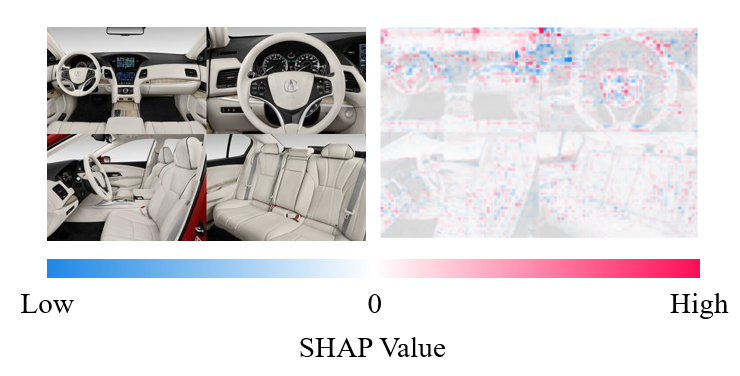}
\caption{SHAP values of the interior image regions of the 2020 Acura RLX\protect\footnotemark[11] for the interior score prediction: the SHAP values on the right with the corresponding interior image on the left. One can observe that most points are clustered near the dashboard, indicating its importance for ratings.}
\label{fig:11}
\end{figure}

To gain further insights into the performance of different features in each region, we used the SHAP method to analyze the image data of individual samples. We exemplify the SHAP values of two representative images for predicting the total and interior scores in Figure~\ref{fig:10} and Figure~\ref{fig:11}, respectively. The red regions positively influence the predictions, while the blue regions negatively influence the predictions. The color intensity indicates the influence extent.

Figure~\ref{fig:10} showcases the SHAP values of the exterior image regions of the 2020 GMC Terrain for predicting the total score. We find that in the front view and angular front view, the regions on the front wheels, the vehicle brand logo, the fog lamps, the front bumper, and the front fenders have positive influences on the total score prediction of this vehicle. Similarly, in the rearview, the regions on the rear wheels, taillights, and bumpers also play a bigger role in predicting the total score. This is reasonable. For example, during night driving, turning on the taillight can alert the following vehicle to maintain a safe distance, and a well-designed bumper can offer better protection in case of accidents. 
This will undoubtedly have a positive impact on the overall rating of the vehicle. Additionally, our SHAP analyses for predicting the other scores show similar trends. However, it is important to note that different vehicles may have distinct components that contribute to different score predictions, thus necessitating a case-by-case analysis by designers and engineers.

Figure~\ref{fig:11} displays the SHAP values of the interior image regions of the 2020 Acura RLX for predicting the interior score. The instrument panel, the gearshift, the dashboard, the steering wheel, the steering wheel controls, and the front and rear seats are likely to have positive impacts on the interior score prediction of this vehicle. Among these, the steering wheel and dashboard have the most substantial impacts, as they are among the most used interior components. We believe that people can only truly experience the comfort and convenience of the front and rear seats when they are in the vehicle, rather than from a picture, so their effect may be weaker than that of the steering wheel. Overall, considering these details and features can lead to a better interior score and increase the vehicle's appeal to potential consumers.

\footnotetext[10]{\url{https://cars.usnews.com/cars-trucks/gmc/terrain/2020/photos-exterior}} 
\footnotetext[11]{\url{https://cars.usnews.com/cars-trucks/acura/rlx/photos-interior}} 

\begin{figure}[h!]
\centering\includegraphics[width=\linewidth]{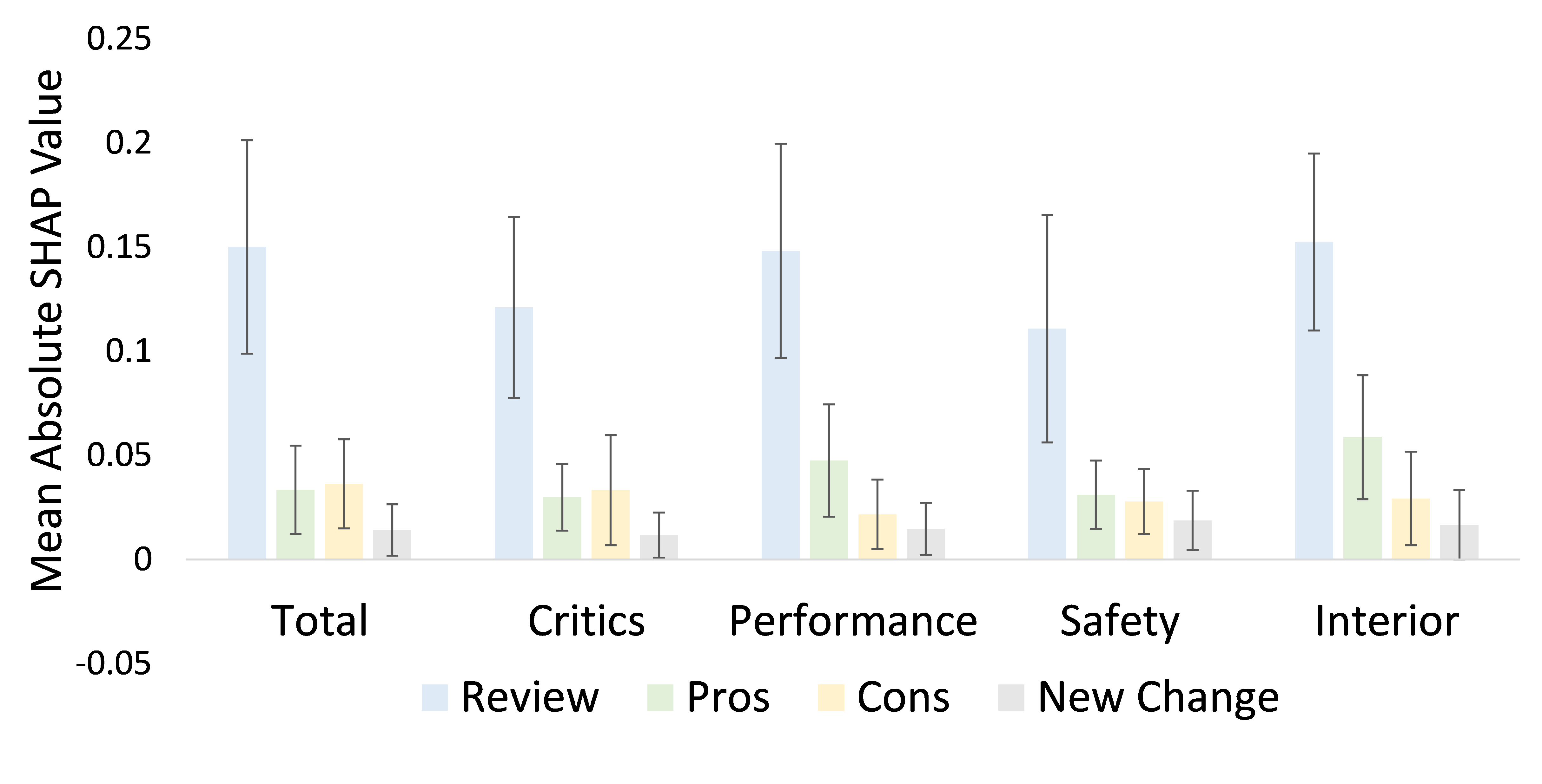}
\caption{Mean Absolute SHAP values of different segments of the text data for predicting different scores with one standard error. The review segment is most important for all predictions.}
\label{fig:12}
\end{figure}

Lastly, We use the SHAP method to analyze the informativeness of different text segments for predicting different rating scores. The results are displayed in Figure~\ref{fig:12}. Our findings reveal that the review of a vehicle and its advantages and disadvantages significantly influence the model predictions. In general, the review segment of the text has the largest influence on the rating score predictions. If this segment provides a negative evaluation (e.g., ) of the vehicle, often with words like ``bottom'', ``however'', or ``but'', the corresponding SHAP value is mostly negative, indicating a negative impact on the rating prediction. Otherwise, words like ``top'' indicate a positive evaluation and tend to have positive impacts on the predictions. Moreover, the brand and year of the vehicle mentioned in the text also have relatively important impacts on the model predictions. Additionally, the pros segment usually has a positive SHAP value, while the cons segment has a negative SHAP value, as expected. The ``New Change'' segment of the text indicates if there are any new changes in the vehicle compared to the previous year. If there are positive changes, the corresponding SHAP value is usually positive.

\begin{figure}[h!]
\centering\includegraphics[width=\linewidth]{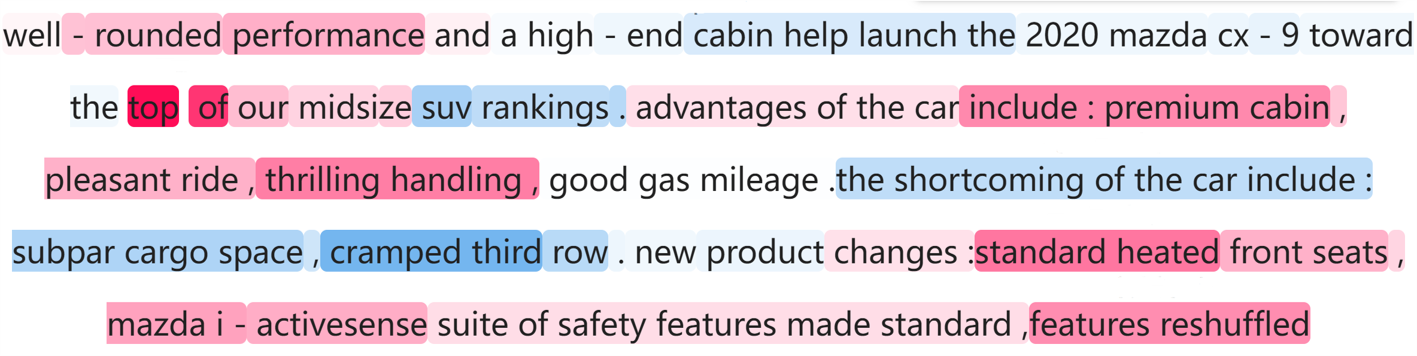}
\caption[Caption for LOF]{SHAP values of the words in the 2020 Mazda CX-9's text data\protect\footnotemark[12] for its total score prediction. The red and blue colors imply positive and negative impacts, respectively. Words such as premium cabin and thrilling handling are found to have high importance.}
\label{fig:13}
\end{figure}

Figure~\ref{fig:13} is an instance of SHAP analysis applied to an individual vehicle - the 2020 Mazda - for predicting its total score. The light blue colors assigned to ``2020'' and ``Mazda'' indicate slightly negative effects on the prediction. The positive SHAP value of the word ``top'' confirms the esteemed position held by the vehicle in the US News evaluation system, positively influencing the prediction. This positive reputation may sway potential buyers towards considering this vehicle over others within this system. In addition, the vehicle has several advantages such as a ``premium cabin,'' ``pleasant ride,'' and ``thrilling handling,'' all of which are positive aspects of the vehicle and present positive SHAP values. These aspects probably lead to a positive perception of the vehicle and attract potential buyers. However, the vehicle also has some drawbacks, such as ``subpar cargo space'' and a ``cramped third row,'' indicating that it may not be the best option for those looking for more space. These downsides have negative impacts on the total score forecast, as indicated by their negative SHAP values. New changes such as ``standard heated front seats'' and ``Mazda i-active sense suite of safety features made standard,'' are positive changes and exhibit positive SHAP values. Engineers and designers need to analyze their own designs case by case.

\footnotetext[12]{\url{https://cars.usnews.com/cars-trucks/mazda/cx-9/2020}} 

To inform the design and optimization directions of individual vehicles, brands, or other aspects, designers and engineers need to extract a suitable sub-dataset from the entire dataset, retrain the prediction model, and carry out SHAP analyses and discussions accordingly. This approach can improve the rating score prediction for a particular type of vehicle and improve the interpretability of the model.

\subsection{Limitations and Future Work}
This section summarizes the limitations of this study and future research directions of using multi-modal learning models to promote engineering design. First, a major limitation of this study is our dataset is much smaller than the other datasets for training large deep learning models, which may not provide sufficient information for the multi-modal learning models to learn the complex interactions between different data modalities. It is hard to harness the full potential of multi-modal learning with small datasets. We observe that the US News website only provides information regarding the vehicles on the US market, leading to the exclusion of some vehicle brands from China, India, and other countries in this study. We will work to expand this dataset by including more vehicle brands and completing the information on the vehicles with missing data items in the future. Second, we use the simple concatenation mechanism to fuse information from different data modalities in this paper, which may lead to less effective information fusion compared to more advanced information fusion mechanisms, such as attention-based or transformer-based information fusion. In future work, we will explore advanced techniques to fuse features learned from the parametric, text, and image data. Additionally, not all information available from the US News website is leveraged in this study. For example, we only select four exterior photos and four interior photos from a much larger photo collection and use a small part of each text description from the website for the rating score prediction. A more comprehensive understanding of a vehicle and a better rating score prediction may be achieved by incorporating all available information into ML. In the future, we need to explore more effective and efficient deep-learning models to manage richer data.


\section{Conclusion}
In this research, we have developed and validated a multi-modal learning model aimed at predicting five different vehicle rating scores—total score, critics score, performance score, safety score, and interior score. 
These predictions are facilitated using the parametric specifications, text descriptions, and images of vehicles. As the foundation of the multi-modal learning model, we developed three unimodal models to independently extract features from parametric, text, and image data. Based on this, we compared the efficacy of the multi-modal learning model against its unimodal equivalents. Our research has led to three significant discoveries: 1. Parametric data proves to be the most informative in predicting all the scores, with the text model surpassing the image model in most instances for predicting the rating scores. 2. The multi-modal learning model, which concurrently learns from parametric, text, and image data, outperforms all the unimodal models. This suggests that multi-modal data learning captures a richer array of information than learning from a single data mode for the task of prediction. 3. The sensitivity analyses conducted via SHAP can offer invaluable insights for interpreting predictions and provide crucial design, optimization, and improvement guidance to designers and engineers. Furthermore, the proposed multi-modal learning methodology can be extrapolated to a broader range of application scenarios, potentially providing fresh insights and inspiration for designers.




\nocite{*}

\bibliographystyle{asmeconf}  
\bibliography{arxiv}

\begin{thebibliography}{10}
\newcommand{\enquote}[1]{``#1''}
\providecommand{\url}[1]{\texttt{#1}}
\providecommand{\urlprefix}{URL }
\expandafter\ifx\csname urlstyle\endcsname\relax
  \providecommand{\doi}[1]{DOI \discretionary{}{}{}#1}\else
  \providecommand{\doi}{DOI \discretionary{}{}{}\begingroup
  \urlstyle{rm}\Url}\fi
\providecommand{\eprint}[2][]{\urlprefix\url{#1#2}}

\bibitem{Simbolon2020TheIO}
Simbolon, Freddy~Pandapotan, Handayani, Elvira and Nugraedy, Menik.
\newblock \enquote{The Influence of Product Quality, Price Fairness, Brand
  Image, and Customer Value on Purchase Decision of Toyota Agya Consumers: A
  Study of Low Cost Green Car.}
\newblock \textit{Binus Business Review} Vol.~11 (2020): pp. 187--196.

\bibitem{Ponmalar2022ReviewApproaches}
Ponmalar, Punitha, Associate, P, Angelin, M~Rs and Assistant, Christinal~C.
\newblock \enquote{{Review on the Pre-owned Car Price Determination using
  Machine Learning Approaches; Review on the Pre-owned Car Price Determination
  using Machine Learning Approaches}.}
  (2022)\doi{10.1109/ICAISS55157.2022.10010958}.

\bibitem{9696839}
Jin, Chuyang.
\newblock \enquote{Price Prediction of Used Cars Using Machine Learning.}
\newblock \textit{2021 IEEE International Conference on Emergency Science and
  Information Technology (ICESIT)}: pp. 223--230. 2021.
\newblock \doi{10.1109/ICESIT53460.2021.9696839}.

\bibitem{tsagris2022advanced}
Tsagris, Michail and Fafalios, Stefanos.
\newblock \enquote{Advanced Car Price Modelling and Prediction.}
\newblock \textit{Advances in Econometrics, Operational Research, Data Science
  and Actuarial Studies: Techniques and Theories}.
\newblock Springer (2022): pp. 479--494.

\bibitem{Xia123ForeXGBoost:XGBoost}
Xia, Zhenchang, Xue, Shan, Wu, ·~Libing, Sun, Jiaxin, Chen, Yanjiao and Zhang,
  Rui.
\newblock \enquote{{ForeXGBoost: passenger car sales prediction based on
  XGBoost}.}
\newblock \textit{Distributed and Parallel Databases} Vol.~38 (123).
\newblock \doi{10.1007/s10619-020-07294-y}.
\newblock \urlprefix\url{https://doi.org/10.1007/s10619-020-07294-y}.

\bibitem{KumarPanda2022CarApproach}
Kumar~Panda, Sandeep, Mohapatra, Ramesh~Kumar, Panda, Subhrakanta, Balamurugan,
  S and Kumar~Panda, Samdeep.
\newblock \enquote{{Car Buying Criteria Evaluation Using Machine Learning
  Approach}.}  (2022)\doi{10.1002/9781119884392.ch10}.
\newblock
  \urlprefix\url{https://onlinelibrary.wiley.com/doi/10.1002/9781119884392.ch10}.

\bibitem{LiAEvaluation}
Li, Deming, Li, Menggang, Han, Gang and Li, Ting.
\newblock \enquote{{A combined deep learning method for internet car
  evaluation}.}
\newblock \textit{Neural Computing and Applications} Vol.~33 .
\newblock \doi{10.1007/s00521-020-05291-x}.
\newblock \urlprefix\url{https://doi.org/10.1007/s00521-020-05291-x}.

\bibitem{Wang2020ScienceDirectLearning}
Wang, Hui~Dong.
\newblock \enquote{{ScienceDirect Research on the Features of Car Insurance
  Data Based on Machine Learning}.}
\newblock \textit{Procedia Computer Science} Vol. 166 (2020): pp. 582--587.
\newblock \doi{10.1016/j.procs.2020.02.016}.
\newblock \urlprefix\url{www.sciencedirect.comwww.sciencedirect.com}.

\bibitem{borisov2022deep}
Borisov, Vadim, Leemann, Tobias, Se{\ss}ler, Kathrin, Haug, Johannes,
  Pawelczyk, Martin and Kasneci, Gjergji.
\newblock \enquote{Deep neural networks and tabular data: A survey.}
\newblock \textit{IEEE Transactions on Neural Networks and Learning Systems}
  (2022).

\bibitem{Su2012LinearRegression}
Su, Xiaogang, Yan, Xin and Tsai, Chih~Ling.
\newblock \enquote{{Linear regression}.}
\newblock \textit{Wiley Interdisciplinary Reviews: Computational Statistics}
  Vol.~4 No.~3 (2012): pp. 275--294.
\newblock \doi{10.1002/WICS.1198}.
\newblock
  \urlprefix\url{https://onlinelibrary.wiley.com/doi/full/10.1002/wics.1198
  https://onlinelibrary.wiley.com/doi/abs/10.1002/wics.1198
  https://wires.onlinelibrary.wiley.com/doi/10.1002/wics.1198}.

\bibitem{wang2005gaussian}
Wang, Jack, Hertzmann, Aaron and Fleet, David~J.
\newblock \enquote{Gaussian process dynamical models.}
\newblock \textit{Advances in neural information processing systems} Vol.~18
  (2005).

\bibitem{chen2016xgboost}
Chen, Tianqi and Guestrin, Carlos.
\newblock \enquote{Xgboost: A scalable tree boosting system.}
\newblock \textit{Proceedings of the 22nd acm sigkdd international conference
  on knowledge discovery and data mining}: pp. 785--794. 2016.

\bibitem{kadra2021regularization}
Kadra, Arlind, Lindauer, Marius, Hutter, Frank and Grabocka, Josif.
\newblock \enquote{Regularization is all you need: Simple neural nets can excel
  on tabular data.}
\newblock \textit{arXiv preprint arXiv:2106.11189} Vol. 536 (2021).

\bibitem{arik2021tabnet}
Arik, Sercan~{\"O} and Pfister, Tomas.
\newblock \enquote{Tabnet: Attentive interpretable tabular learning.}
\newblock \textit{Proceedings of the AAAI Conference on Artificial
  Intelligence}, Vol.~35. ~8: pp. 6679--6687. 2021.

\bibitem{vaswani2017attention}
Vaswani, Ashish, Shazeer, Noam, Parmar, Niki, Uszkoreit, Jakob, Jones, Llion,
  Gomez, Aidan~N, Kaiser, {\L}ukasz and Polosukhin, Illia.
\newblock \enquote{Attention is all you need.}
\newblock \textit{Advances in neural information processing systems} Vol.~30
  (2017).

\bibitem{8320684}
Pak, Myeongsuk and Kim, Sanghoon.
\newblock \enquote{A review of deep learning in image recognition.}
\newblock \textit{2017 4th International Conference on Computer Applications
  and Information Processing Technology (CAIPT)}: pp. 1--3. 2017.
\newblock \doi{10.1109/CAIPT.2017.8320684}.

\bibitem{rawat2017deep}
Rawat, Waseem and Wang, Zenghui.
\newblock \enquote{Deep convolutional neural networks for image classification:
  A comprehensive review.}
\newblock \textit{Neural computation} Vol.~29 No.~9 (2017): pp. 2352--2449.

\bibitem{minaee2021image}
Minaee, Shervin, Boykov, Yuri, Porikli, Fatih, Plaza, Antonio, Kehtarnavaz,
  Nasser and Terzopoulos, Demetri.
\newblock \enquote{Image segmentation using deep learning: A survey.}
\newblock \textit{IEEE transactions on pattern analysis and machine
  intelligence} Vol.~44 No.~7 (2021): pp. 3523--3542.

\bibitem{elasri2022image}
Elasri, Mohamed, Elharrouss, Omar, Al-Maadeed, Somaya and Tairi, Hamid.
\newblock \enquote{Image Generation: A Review.}
\newblock \textit{Neural Processing Letters} Vol.~54 No.~5 (2022): pp.
  4609--4646.

\bibitem{krizhevsky2017imagenet}
Krizhevsky, Alex, Sutskever, Ilya and Hinton, Geoffrey~E.
\newblock \enquote{Imagenet classification with deep convolutional neural
  networks.}
\newblock \textit{Communications of the ACM} Vol.~60 No.~6 (2017): pp. 84--90.

\bibitem{simonyan2014very}
Simonyan, Karen and Zisserman, Andrew.
\newblock \enquote{Very deep convolutional networks for large-scale image
  recognition.}
\newblock \textit{arXiv preprint arXiv:1409.1556}  (2014).

\bibitem{he2016deep}
He, Kaiming, Zhang, Xiangyu, Ren, Shaoqing and Sun, Jian.
\newblock \enquote{Deep residual learning for image recognition.}
\newblock \textit{Proceedings of the IEEE conference on computer vision and
  pattern recognition}: pp. 770--778. 2016.

\bibitem{szegedy2015going}
Szegedy, Christian, Liu, Wei, Jia, Yangqing, Sermanet, Pierre, Reed, Scott,
  Anguelov, Dragomir, Erhan, Dumitru, Vanhoucke, Vincent and Rabinovich,
  Andrew.
\newblock \enquote{Going deeper with convolutions.}
\newblock \textit{Proceedings of the IEEE conference on computer vision and
  pattern recognition}: pp. 1--9. 2015.

\bibitem{bengio2000neural}
Bengio, Yoshua, Ducharme, R{\'e}jean and Vincent, Pascal.
\newblock \enquote{A neural probabilistic language model.}
\newblock \textit{Advances in neural information processing systems} Vol.~13
  (2000).

\bibitem{schmidhuber2015deep}
Schmidhuber, J{\"u}rgen.
\newblock \enquote{Deep learning in neural networks: An overview.}
\newblock \textit{Neural networks} Vol.~61 (2015): pp. 85--117.

\bibitem{hochreiter1997long}
Hochreiter, Sepp and Schmidhuber, J{\"u}rgen.
\newblock \enquote{Long short-term memory.}
\newblock \textit{Neural computation} Vol.~9 No.~8 (1997): pp. 1735--1780.

\bibitem{chung2014empirical}
Chung, Junyoung, Gulcehre, Caglar, Cho, KyungHyun and Bengio, Yoshua.
\newblock \enquote{Empirical evaluation of gated recurrent neural networks on
  sequence modeling.}
\newblock \textit{arXiv preprint arXiv:1412.3555}  (2014).

\bibitem{radford2018improving}
Radford, Alec, Narasimhan, Karthik, Salimans, Tim, Sutskever, Ilya et~al.
\newblock \enquote{Improving language understanding by generative
  pre-training.}  (2018).

\bibitem{radford2019language}
Radford, Alec, Wu, Jeffrey, Child, Rewon, Luan, David, Amodei, Dario,
  Sutskever, Ilya et~al.
\newblock \enquote{Language models are unsupervised multitask learners.}
\newblock \textit{OpenAI blog} Vol.~1 No.~8 (2019): p.~9.

\bibitem{devlin2018bert}
Devlin, Jacob, Chang, Ming-Wei, Lee, Kenton and Toutanova, Kristina.
\newblock \enquote{Bert: Pre-training of deep bidirectional transformers for
  language understanding.}
\newblock \textit{arXiv preprint arXiv:1810.04805}  (2018).

\bibitem{liu2019roberta}
Liu, Yinhan, Ott, Myle, Goyal, Naman, Du, Jingfei, Joshi, Mandar, Chen, Danqi,
  Levy, Omer, Lewis, Mike, Zettlemoyer, Luke and Stoyanov, Veselin.
\newblock \enquote{Roberta: A robustly optimized bert pretraining approach.}
\newblock \textit{arXiv preprint arXiv:1907.11692}  (2019).

\bibitem{yang2019xlnet}
Yang, Zhilin, Dai, Zihang, Yang, Yiming, Carbonell, Jaime, Salakhutdinov,
  Russ~R and Le, Quoc~V.
\newblock \enquote{Xlnet: Generalized autoregressive pretraining for language
  understanding.}
\newblock \textit{Advances in neural information processing systems} Vol.~32
  (2019).

\bibitem{Song2023Multi-modalDirections}
Song, Binyang, Zhou, Rui and Ahmed, Faez.
\newblock \enquote{{Multi-modal Machine Learning in Engineering Design: A
  Review and Future Directions}.}  (2023)\doi{10.48550/arxiv.2302.10909}.
\newblock \urlprefix\url{https://arxiv.org/abs/2302.10909v1}.

\bibitem{Baltrusaitis2019multimodalTaxonomy}
Baltrusaitis, Tadas, Ahuja, Chaitanya and Morency, Louis~Philippe.
\newblock \enquote{{Multimodal Machine Learning: A Survey and Taxonomy}.}
\newblock \textit{IEEE Transactions on Pattern Analysis and Machine
  Intelligence} Vol.~41 No.~2 (2019): pp. 423--443.
\newblock \doi{10.1109/TPAMI.2018.2798607}.

\bibitem{Song2022HEYAnd}
Song, Binyang, Miller, Scarlett and Ahmed, Faez.
\newblock \enquote{Hey, AI! Can You See What I See? Multimodal Transfer
  Learning-Based Design Metrics Prediction for Sketches With Text
  Descriptions.}
\newblock \textit{International Design Engineering Technical Conferences and
  Computers and Information in Engineering Conference}, Vol. 86267: p.
  V006T06A017. 2022. American Society of Mechanical Engineers.

\bibitem{Nojavanasghari2016DeepPrediction}
Nojavanasghari, Behnaz, Gopinath, Deepak, Koushik, Jayanth, Baltru{\v{s}}aitis,
  Tadas and Morency, Louis~Philippe.
\newblock \enquote{{Deep multimodal fusion for persuasiveness prediction}.}
\newblock \textit{ICMI 2016 - Proceedings of the 18th ACM International
  Conference on Multimodal Interaction}  (2016): pp.
  284--288\doi{10.1145/2993148.2993176}.

\bibitem{Anastasopoulos2019NeuralFeatures}
Anastasopoulos, Antonios, Kumar, Shankar and Liao, Hank.
\newblock \enquote{{Neural Language Modeling with Visual Features}.}
\newblock \textit{undefined}
  (2019)\urlprefix\url{http://arxiv.org/abs/1903.02930
  https://arxiv.org/abs/1903.02930v1}.

\bibitem{Vielzeuf2019CentralNet:Fusion}
Vielzeuf, Valentin, Lechervy, Alexis, Pateux, Stéphane and Jurie, Frédéric.
\newblock \enquote{{CentralNet: A multilayer approach for multimodal fusion}.}
\newblock \textit{Lecture Notes in Computer Science (including subseries
  Lecture Notes in Artificial Intelligence and Lecture Notes in
  Bioinformatics)} Vol. 11134 LNCS (2019): pp. 575--589.
\newblock \doi{10.1007/978-3-030-11024-6{\_}44}.

\bibitem{Shutova2016BlackFeatures}
Shutova, Ekaterina, Kiela, Douwe and Maillard, Jean.
\newblock \enquote{{Black Holes and White Rabbits: Metaphor Identification with
  Visual Features}.}
\newblock \textit{2016 Conference of the North American Chapter of the
  Association for Computational Linguistics: Human Language Technologies, NAACL
  HLT 2016 - Proceedings of the Conference}  (2016): pp.
  160--170\doi{10.18653/V1/N16-1020}.
\newblock \urlprefix\url{https://aclanthology.org/N16-1020}.

\bibitem{Cao2016DeepRetrieval}
Cao, Yue, Long, Mingsheng, Wang, Jianmin, Yang, Qiang and Yuy, Philip~S.
\newblock \enquote{{Deep visual-semantic hashing for cross-modal retrieval}.}
\newblock \textit{Proceedings of the ACM SIGKDD International Conference on
  Knowledge Discovery and Data Mining} Vol. 13-17-Augu (2016): pp. 1445--1454.
\newblock \doi{10.1145/2939672.2939812}.

\bibitem{Morvant2014MajorityFusion}
Morvant, Emilie, Habrard, Amaury and Ayache, Stéphane.
\newblock \enquote{{Majority Vote of Diverse Classifiers for Late Fusion}.}
\newblock \textit{Lecture Notes in Computer Science (including subseries
  Lecture Notes in Artificial Intelligence and Lecture Notes in
  Bioinformatics)} Vol. 8621 LNCS (2014): pp. 153--162.
\newblock \doi{10.48550/arxiv.1404.7796}.
\newblock \urlprefix\url{https://arxiv.org/abs/1404.7796v2}.

\bibitem{Zadeh2017TensorAnalysis}
Zadeh, Amir, Chen, Minghai, Cambria, Erik, Poria, Soujanya and Morency,
  Louis~Philippe.
\newblock \enquote{{Tensor Fusion Network for Multimodal Sentiment Analysis}.}
\newblock \textit{EMNLP 2017 - Conference on Empirical Methods in Natural
  Language Processing, Proceedings}  (2017): pp.
  1103--1114\doi{10.48550/arxiv.1707.07250}.
\newblock \urlprefix\url{https://arxiv.org/abs/1707.07250v1}.

\bibitem{Chen2019PathomicPrognosis}
Chen, Richard~J., Lu, Ming~Y., Wang, Jingwen, Williamson, Drew~F.K., Rodig,
  Scott~J., Lindeman, Neal~I. and Mahmood, Faisal.
\newblock \enquote{{Pathomic Fusion: An Integrated Framework for Fusing
  Histopathology and Genomic Features for Cancer Diagnosis and Prognosis}.}
\newblock \textit{IEEE Transactions on Medical Imaging} Vol.~41 No.~4 (2019):
  pp. 757--770.
\newblock \doi{10.48550/arxiv.1912.08937}.
\newblock \urlprefix\url{https://arxiv.org/abs/1912.08937v3}.

\bibitem{Tenenbaum2000SeparatingModels}
Tenenbaum, Joshua~B. and Freeman, William~T.
\newblock \enquote{{Separating style and content with bilinear models}.}
\newblock \textit{Neural Computation} Vol.~12 No.~6 (2000): pp. 1247--1283.
\newblock \doi{10.1162/089976600300015349}.

\bibitem{Graves2014NeuralMachines}
Graves, Alex, Wayne, Greg and Danihelka, Ivo.
\newblock \enquote{{Neural Turing Machines}.}
\newblock \textit{arXiv preprint arXiv:1410.5401.}
  (2014)\urlprefix\url{https://arxiv.org/abs/1410.5401v2
  http://arxiv.org/abs/1410.5401}.

\bibitem{Rombach2021High-ResolutionModels}
Rombach, Robin, Blattmann, Andreas, Lorenz, Dominik, Esser, Patrick and Ommer,
  Björn.
\newblock \enquote{{High-Resolution Image Synthesis with Latent Diffusion
  Models}.}  (2021): pp. 10674--10685\doi{10.48550/arxiv.2112.10752}.
\newblock \urlprefix\url{https://arxiv.org/abs/2112.10752v2}.

\bibitem{Liu2019Point-VoxelLearning}
Liu, Zhijian, Tang, Haotian, Lin, Yujun and Han, Song.
\newblock \enquote{{Point-Voxel CNN for Efficient 3D Deep Learning}.}
\newblock \textit{Advances in Neural Information Processing Systems} Vol.~32
  (2019).
\newblock \doi{10.48550/arxiv.1907.03739}.
\newblock \urlprefix\url{https://arxiv.org/abs/1907.03739v2}.

\bibitem{Nichol2021GLIDE:Models}
Nichol, Alex, Dhariwal, Prafulla, Ramesh, Aditya, Shyam, Pranav, Mishkin,
  Pamela, McGrew, Bob, Sutskever, Ilya and Chen, Mark.
\newblock \enquote{{GLIDE: Towards Photorealistic Image Generation and Editing
  with Text-Guided Diffusion Models}.}  (2021)\doi{10.48550/arxiv.2112.10741}.
\newblock \urlprefix\url{https://arxiv.org/abs/2112.10741v3}.

\bibitem{Kim2021DiffusionCLIP:Manipulation}
Kim, Gwanghyun, Kwon, Taesung and Ye, Jong~Chul.
\newblock \enquote{{DiffusionCLIP: Text-Guided Diffusion Models for Robust
  Image Manipulation}.}  (2021)\doi{10.48550/arxiv.2110.02711}.
\newblock \urlprefix\url{https://arxiv.org/abs/2110.02711v6}.

\bibitem{DucTuan2021MultimodalDetection}
Duc~Tuan, Nguyen~Manh and Quang Nhat~Minh, Pham.
\newblock \enquote{{Multimodal Fusion with BERT and Attention Mechanism for
  Fake News Detection}.}
\newblock \textit{Proceedings - 2021 RIVF International Conference on Computing
  and Communication Technologies, RIVF 2021}
  (2021)\doi{10.48550/arxiv.2104.11476}.
\newblock \urlprefix\url{https://arxiv.org/abs/2104.11476v2}.

\bibitem{Devlin2015LanguageWorks}
Devlin, Jacob, Cheng, Hao, Fang, Hao, Gupta, Saurabh, Deng, Li, He, Xiaodong,
  Zweig, Geoffrey and Mitchell, Margaret.
\newblock \enquote{{Language Models for Image Captioning: The Quirks and What
  Works}.}
\newblock \textit{ACL-IJCNLP 2015 - 53rd Annual Meeting of the Association for
  Computational Linguistics and the 7th International Joint Conference on
  Natural Language Processing of the Asian Federation of Natural Language
  Processing, Proceedings of the Conference} Vol.~2 (2015): pp. 100--105.
\newblock \doi{10.48550/arxiv.1505.01809}.
\newblock \urlprefix\url{https://arxiv.org/abs/1505.01809v3}.

\bibitem{Kwon2022EnablingLearning}
Kwon, Elisa, Huang, Forrest and Goucher-Lambert, Kosa.
\newblock \enquote{{Enabling multi-modal search for inspirational design
  stimuli using deep learning}.}
\newblock \textit{AI EDAM} Vol.~36 (2022): p. e22.
\newblock \doi{10.1017/S0890060422000130}.
\newblock
  \urlprefix\url{https://www.cambridge.org/core/journals/ai-edam/article/enabling-multimodal-search-for-inspirational-design-stimuli-using-deep-learning/2F5EA4243AD422EA74EA2B9FDAF8FF05}.

\bibitem{Yuan2022LeveragingModel}
Yuan, Chenxi, Marion, Tucker and Moghaddam, Mohsen.
\newblock \enquote{{Leveraging End-User Data for Enhanced Design Concept
  Evaluation: A Multimodal Deep Regression Model}.}
\newblock \textit{Journal of Mechanical Design} Vol. 144 No.~2 (2022): pp.
  1--20.
\newblock \doi{10.1115/1.4052366}.
\newblock
  \urlprefix\url{https://asmedigitalcollection.asme.org/mechanicaldesign/article/144/2/021403/1119449/Leveraging-End-User-Data-for-Enhanced-Design}.

\bibitem{Li2022AAutoencoder}
Li, Xingang, Xie, Charles and Sha, Zhenghui.
\newblock \enquote{{A Predictive and Generative Design Approach for
  Three-Dimensional Mesh Shapes Using Target-Embedding Variational
  Autoencoder}.}
\newblock \textit{Journal of Mechanical Design} Vol. 144 No.~11 (2022).
\newblock \doi{10.1115/1.4054906}.
\newblock
  \urlprefix\url{https://asmedigitalcollection.asme.org/mechanicaldesign/article/144/11/114501/1141958/A-Predictive-and-Generative-Design-Approach-for}.

\bibitem{Song2023ATTENTION-ENHANCEDEVALUATIONS}
Song, Binyang, Miller, Scarlett and Ahmed, Faez.
\newblock \enquote{Attention-enhanced Multimodal Learning For Conceptual Design
  Evaluation.}
\newblock \textit{Journal of Mechanical Design}  (2023): pp.
  1--38\doi{10.1115/1.4056669}.
\newblock
  \urlprefix\url{https://asmedigitalcollection.asme.org/mechanicaldesign/article/doi/10.1115/1.4056669/1156042/ATTENTION-ENHANCED-MULTIMODAL-LEARNING-FOR}.

\bibitem{murdoch2019definitions}
Murdoch, W~James, Singh, Chandan, Kumbier, Karl, Abbasi-Asl, Reza and Yu, Bin.
\newblock \enquote{Definitions, methods, and applications in interpretable
  machine learning.}
\newblock \textit{Proceedings of the National Academy of Sciences} Vol. 116
  No.~44 (2019): pp. 22071--22080.

\bibitem{molnar2020interpretable}
Molnar, Christoph.
\newblock \textit{Interpretable machine learning}.
\newblock Lulu. com (2020).

\bibitem{molnar2021interpretable}
Molnar, Christoph, Casalicchio, Giuseppe and Bischl, Bernd.
\newblock \enquote{Interpretable machine learning--a brief history,
  state-of-the-art and challenges.}
\newblock \textit{ECML PKDD 2020 Workshops: Workshops of the European
  Conference on Machine Learning and Knowledge Discovery in Databases (ECML
  PKDD 2020): SoGood 2020, PDFL 2020, MLCS 2020, NFMCP 2020, DINA 2020, EDML
  2020, XKDD 2020 and INRA 2020, Ghent, Belgium, September 14--18, 2020,
  Proceedings}: pp. 417--431. 2021. Springer.

\bibitem{fisher2019all}
Fisher, Aaron, Rudin, Cynthia and Dominici, Francesca.
\newblock \enquote{All Models are Wrong, but Many are Useful: Learning a
  Variable's Importance by Studying an Entire Class of Prediction Models
  Simultaneously.}
\newblock \textit{J. Mach. Learn. Res.} Vol.~20 No. 177 (2019): pp. 1--81.

\bibitem{rose2019explanatory}
Ros{\'e}, Carolyn~P, McLaughlin, Elizabeth~A, Liu, Ran and Koedinger,
  Kenneth~R.
\newblock \enquote{Explanatory learner models: Why machine learning (alone) is
  not the answer.}
\newblock \textit{British Journal of Educational Technology} Vol.~50 No.~6
  (2019): pp. 2943--2958.

\bibitem{ahmed2022product}
Ahmed, Faez, Cui, Yaxin, Fu, Yan and Chen, Wei.
\newblock \enquote{Product Competition Prediction in Engineering Design Using
  Graph Neural Networks.}
\newblock \textit{ASME Open Journal of Engineering} Vol.~1 (2022).

\bibitem{shrikumar2017learning}
Shrikumar, Avanti, Greenside, Peyton and Kundaje, Anshul.
\newblock \enquote{Learning important features through propagating activation
  differences.}
\newblock \textit{International conference on machine learning}: pp.
  3145--3153. 2017. PMLR.

\bibitem{sundararajan2020many}
Sundararajan, Mukund and Najmi, Amir.
\newblock \enquote{The many Shapley values for model explanation.}
\newblock \textit{International conference on machine learning}: pp.
  9269--9278. 2020. PMLR.

\bibitem{NIPS2017_7062}
Lundberg, Scott~M and Lee, Su-In.
\newblock \enquote{A Unified Approach to Interpreting Model Predictions.}
\newblock Guyon, I., Luxburg, U.~V., Bengio, S., Wallach, H., Fergus, R.,
  Vishwanathan, S. and Garnett, R. (eds.). \textit{Advances in Neural
  Information Processing Systems 30}.
\newblock Curran Associates, Inc. (2017): pp. 4765--4774.
\newblock
  \urlprefix\url{http://papers.nips.cc/paper/7062-a-unified-approach-to-interpreting-model-predictions.pdf}.

\bibitem{sundararajan2017axiomatic}
Sundararajan, Mukund, Taly, Ankur and Yan, Qiqi.
\newblock \enquote{Axiomatic attribution for deep networks.}
\newblock \textit{International conference on machine learning}: pp.
  3319--3328. 2017. PMLR.

\bibitem{smilkov2017smoothgrad}
Smilkov, Daniel, Thorat, Nikhil, Kim, Been, Vi{\'e}gas, Fernanda and
  Wattenberg, Martin.
\newblock \enquote{Smoothgrad: removing noise by adding noise.}
\newblock \textit{arXiv preprint arXiv:1706.03825}  (2017).

\end{thebibliography}

\end{document}